\documentclass[runningheads]{llncs}

\usepackage{eccv}

\usepackage{eccvabbrv}

\usepackage{graphicx}
\usepackage{booktabs}

\usepackage[accsupp]{axessibility}  

\usepackage{hyperref}

\usepackage{orcidlink}

\usepackage[table]{xcolor}
\usepackage{multirow}
\usepackage{siunitx}
\usepackage{array}
\usepackage{caption}
\usepackage{rotating}
\usepackage{adjustbox}
\usepackage[utf8]{inputenc}
\usepackage{amsmath}
\usepackage{colortbl}
\usepackage{makecell} 
\usepackage{arydshln}
\usepackage{paralist}

\newcommand*{\ourmodel}{EvDiff\@\xspace}
\newcommand*{\ourencoder}{EvEncoder\@\xspace}

\usepackage{tikz}       
\newcommand*\blackcircled[1]{\tikz[baseline=(char.base)]{
            \node[shape=circle,draw,inner sep=0pt,fill=black,text=white] (char) {#1};}}
\newcommand{\myhyperlink}[3][black]{\hyperlink{#2}{\color{#1}{#3}}} 
\newcommand{\encoder}{\mathcal{E}_{event}}

\newcommand{\pub}[1]{{\color{gray}{\tiny{[{#1}]\!}}}}
\newcommand{\best}[1]{\textcolor{red}{#1}}
\newcommand{\second}{\color{blue}}
\definecolor{tableHeadGray}{rgb}{0.9, 0.9, 0.9}  
\newcommand{\customfootnotetext}[1]{{\let\thefootnote\relax\footnotetext{#1}}}

\begin{document}

\title{EvDiff: Event-Based Video Reconstruction using One-Step Diffusion Models}
\titlerunning{EvDiff}

\author{Weilun Li\inst{1,2,*} \and
    Lei Sun\inst{2,*\dagger} \and
    Ruixi Gao\inst{1} \and
    Qi Jiang\inst{1} \and
    Yuqin Ma\inst{1} \and
    Kaiwei Wang\inst{1\dagger} \and
    Ming-Hsuan Yang\inst{3,4} \and
    Luc Van Gool\inst{2} \and
    Danda Pani Paudel\inst{2}
}

\authorrunning{W. Li, L. Sun \textit{et al}.}


\institute{Zhejiang University, Hangzhou, China \and
INSAIT, Sofia University ``St. Kliment Ohridski'', Sofia, Bulgaria \and
UC Merced, CA, USA \and
Google DeepMind, London, United Kingdom \\}

\maketitle
\customfootnotetext{$^*$ Equal contribution. Work mainly done during Weilun Li's internship at INSAIT.}
\customfootnotetext{$^\dagger$ Corresponding authors: \email{leo\_sun@zju.edu.cn}, \email{wangkaiwei@zju.edu.cn}}

\begin{center}
    \centering
    \captionsetup{type=figure}
    \includegraphics[width=\linewidth]{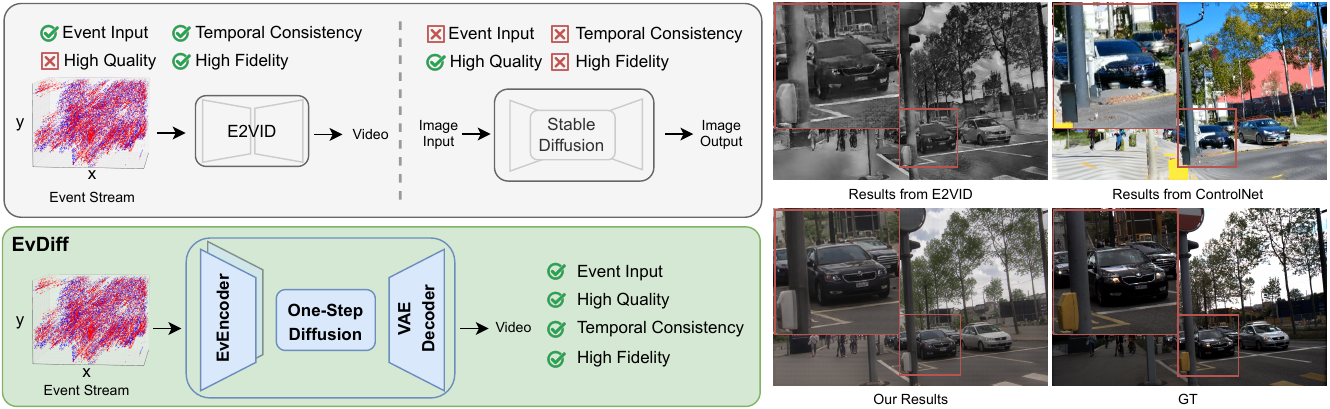}
    \captionof{figure}{Our EvDiff can reconstruct real high-quality video streams from monochrome event streams, while maintaining both fidelity and realism. Compared with Ground-Truth (GT), our result shows a higher dynamic range. }
    \label{fig:first_page_fig}
\end{center}

\begin{abstract}
As neuromorphic sensors, event cameras asynchronously record changes in brightness as streams of sparse events with the advantages of high temporal resolution and high dynamic range.  
Reconstructing intensity images from events is a highly ill-posed task due to the inherent ambiguity of absolute brightness.
Early methods generally follow an end-to-end regression paradigm, directly mapping events to intensity frames in a deterministic manner. While effective to some extent, these approaches often yield perceptually inferior results and struggle to scale up in model capacity and training data.
In this work, we propose EvDiff, an event-based diffusion model that follows a surrogate training framework to produce high-quality videos.
To reduce the high computational cost of high-frame-rate video generation, we design an event-based diffusion model that performs only a single forward diffusion step, equipped with a temporally consistent EvEncoder.
Furthermore, our novel Surrogate Training Framework eliminates the dependence on paired event–image datasets, allowing the model to leverage large-scale image datasets for higher capacity.
The proposed EvDiff is capable of generating high-quality colorful videos solely from monochromatic event streams.
Experiments on real-world datasets demonstrate that our method strikes a sweet spot between fidelity and realism, outperforming existing approaches on both pixel-level and perceptual metrics. The code will be released publicly.
%
\end{abstract}

\section{Introduction}
\label{sec:intro}

Inspired by the human visual system, the Silicon Retina~\cite{mahowald1994silicon} introduced the foundation of perceptual sensing with neuromorphic cameras, commonly referred to as Dynamic Vision Sensors (DVS) or event cameras. Unlike conventional frame-based sensors, event cameras asynchronously record local changes in light intensity as streams of discrete events. Owing to the sparse nature, they provide unique advantages over their frame-based counterparts, including low power consumption, low latency, microsecond-level temporal resolution, and an ultra-high dynamic range (HDR) of up to $140$ dB~\cite{gallego2020event,patrick2008128x}.

However, due to their sparse and asynchronous nature, raw event streams cannot be directly processed by modern computer vision architectures unless specialized models such as spiking neural networks (SNNs)~\cite{tavanaei2019deep} are employed. To mitigate this gap, researchers often convert event streams into middle representations, such as voxel grids~\cite{zhu2018ev} or time surfaces~\cite{manderscheid2019speed}. A more direct and intuitive approach is to reconstruct video frames from events, which not only enables the visualization of high-speed and HDR videos but also bridges event cameras to the vast ecosystem of frame-based computer vision methods~\cite{rebecq2019high,rebecq2019events}. 

While early works~\cite{bardow2016simultaneous, munda2018real, scheerlinck2018continuous} incorporated handcrafted smoothness priors into video reconstruction, a major breakthrough was achieved by E2VID~\cite{rebecq2019events}, which introduced a data-driven learning approach to event-to-video conversion. Subsequent studies have refined this paradigm by simplifying the network architecture~\cite{2020FireNet}, adding self-supervised constraints~\cite{SSL-E2VID}, and explicitly modeling the sparsity of events~\cite{cadena2023sparse}, among others. Nevertheless, most existing approaches remain constrained by the paradigm introduced by E2VID, which relies on relatively small event–image paired datasets and compact network architectures. Owing to the scarcity of large-scale training data and the inherently ill-posed nature of the reconstruction problem, these methods typically yield low-quality grayscale outputs that fall short of perceptual satisfaction for human observers.

Meanwhile, recent advances in generative models have been largely driven by the success of diffusion models~\cite{rombach2022high,ho2020denoising,diffusionsurvey}. One of the key factors behind their effectiveness lies in their scaling ability, which allows them to fully exploit large model capacities and massive high-quality datasets~\cite{schuhmann2022laion,places365}.

Motivated by this trend, bringing the power of large diffusion models to event-based vision naturally becomes an appealing goal. However, two major obstacles stand in the way: 
\hypertarget{P1}{\blackcircled{1}}~\textit{Scaling up the model size} to diffusion-based video generation frameworks is \textit{challenging} due to the prohibitively high computational costs, as event data are inherently high-speed and the corresponding videos must be rendered at high frame rates to preserve this temporal fidelity.
\hypertarget{P2}{\blackcircled{2}} Compared to contemporary large-scale image datasets, \textit{event–image paired datasets are orders of magnitude smaller}, and their large-scale collection is nearly impossible due to the fact that event cameras are not yet widely adopted. Simulators~\cite{rebecq2018esim,hu2021v2e} also require high-frame-rate videos as inputs, which are limited in availability.

In this work, aiming to bring the powerful large foundational model to event-to-video reconstruction, we introduce \ourmodel. At the model level, we adopt the Stable Diffusion 3 (SD3) model~\cite{esser2024scaling} as our base model. To address the problem of overwhelming computation burden to produce high-framerate video from events, we propose \ourencoder to encode events and temporal information into latent presentations, following a one-step diffusion model with SD3 as base model. 
At the data level, to leverage the large-scale data to feed the model, we design a Surrogate Training Pipeline to use the low-quality coarse reconstruction results to bridge the event data and image data. 
To be more specific, first, we design an E2VID-style Degradation Model to synthesize E2VID results from high-quality images, and we train our one-step EvDiff with the synthesized E2VID results (coarse results) and their high-quality counterparts in Place 365~\cite{places365}, a large-scale image dataset with $1.8$ million images. Next, we distill the E2VID and VAE encoder to our \ourencoder and then we finetune the whole model. By utilizing the Surrogate Training Pipeline, we extend the training data from the event-image-paired dataset to a much broader scale. Experiments on real-world datasets are conducted, and our \ourmodel produces higher-quality chromatic videos than regression-based methods. As shown in Fig.~\ref{fig:first_page_fig}, compared with ControlNet-based counterparts~\cite{zhang2023adding}, our method is more efficient, exhibits stronger temporal consistency, and produces higher-fidelity results without relying on prompts derived from ground-truth images. Moreover, even when compared directly with the ground-truth images, our reconstructed videos often display a higher dynamic range, owing to the inherent sensing characteristics of event cameras.

The main contributions of this work are:

\begin{compactitem}
\item We revisit the event-based video reconstruction, and propose to transfer the capabilities of large-scale diffusion models to event-based video reconstruction, achieving high reconstruction fidelity and realism while maintaining efficient inference.

\item We design a novel Surrogate Training Pipeline along with the E2VID-style Degradation Model that serves as a bridge between scarce event-video paired data and large-scale natural image datasets, enabling effective model training at scale.   

\item Experiments show that our method reconstructs faithful and chromatic videos from monochromatic events only, competing favorably against state-of-the-art methods and the ControlNet-based counterpart.
\end{compactitem}
\section{Related Work}
\label{sec:related_work}

\subsection{Event-Based Video Reconstruction}
\label{subsec:e2vid}

Due to their unique advantages, such as high temporal resolution, low latency, and robustness under challenging illumination, event cameras have found applications across a wide range of vision tasks, including SLAM, feature tracking, image deblurring~\cite{vitoria2022event,sun2021mefnet,EventGuidedDeblurring,jiang2020learning,xu2021motion_deblur_events,sun2025ntire,sun2022event,sun2026second,EventGuidedDeblurring,FrequencyAwareEventBasedDeblurring,CMTA,TowardsRealWorldDeblurring}, low-light enhancement~\cite{sun2025low,EventguidedUnifiedFrameworkInterpolation}, video interpolation~\cite{tulyakov2021time,he2022timereplayer,tulyakov2022time,zhang2022unifying,lin2020learning_event_video_deblur,sun2023event,edi_pan,haoyu2020learning,sun2024unified,EventBasedInterpolationCrossModal}, and other tasks~\cite{ye2025towards,lu2024sge,bao2023improving,chen2022efficient,Cho2024ABD}. In addition to these tasks, we investigate the use of event cameras for video reconstruction~\cite{bardow2016simultaneous, scheerlinck2018continuous, munda2018real, rebecq2019events, rebecq2019high, wang2019event, 2020FireNet, SSL-E2VID, han2021evintsr, cadena2023sparse}, \ie, recovering intensity frames from event streams, in this work.

Early approaches~\cite{bardow2016simultaneous,scheerlinck2018continuous,munda2018real} relied on carefully designed physical models with integration or filtering techniques, often requiring additional prior information. With the advent of deep learning, reconstruction quality improved significantly. A milestone was E2VID~\cite{rebecq2019events,rebecq2019high}, which combined a U-Net with temporal modules and achieved strong results at moderate computational cost. Later works such as FireNet~\cite{2020FireNet} explored lightweight architectures, while E2VID+ and FireNet+~\cite{E2VID+} addressed the shortage of training data by introducing synthetic events generated with ESIM~\cite{rebecq2018esim}, a strategy widely adopted in subsequent research. Following E2VID-style variants (\eg, SPADE-E2VID~\cite{Cadena2021Spade-E2VID}, Hyper-E2VID~\cite{ercan2024hypere2vid}, SCSE-E2VID~\cite{Lu2022SCSE-E2VID}, Sparse-E2VID~\cite{cadena2023sparse}) introduced different architectural modules, typically yielding incremental improvements. Other works have explored event-based methods in resource-constrained scenarios, either using spiking neural networks or self-supervised learning~\cite{SSL-E2VID} paradigms. The aforementioned works follow the same end-to-end training paradigm as E2VID, which typically results in relatively low-quality grayscale reconstructions with visually noticeable artifacts.
Recent endeavors have sought to move beyond this paradigm. For example, diffusion-based E2VIDiff~\cite{liang2024e2vidiff} proposes to condition Stable Diffusion~\cite{rombach2022high} on events for the first time, producing chromatic reconstructions but with limited fidelity, leading to substantial deviations from the ground truth (GT).

\subsection{One-Step Diffusion Models}
\label{subsec:diff}

Diffusion models~\cite{ho2020denoising,sohl2015deep,song2019generative,song2020score,qian2026learning,su2026bridging}, which generate data by iteratively denoising random noise through a learned stochastic process, have recently achieved remarkable success as state-of-the-art deep generative models~\cite{diffusionsurvey}. When scaled up with billions of parameters and trained on massive natural image datasets~\cite{schuhmann2022laion,wang2025lavie}, they demonstrate impressive realism and strong generative priors, enabling high-quality image and video synthesis~\cite{rombach2022high,blattmann2023stable,ramesh2021zero,saharia2022photorealistic,wan2025wan}. However, these benefits come at the cost of heavy computational requirements, as classical diffusion processes involve hundreds of iterative denoising steps~\cite{ho2020denoising,ddim}, which limit their practicality in time-sensitive and high frame-rate video applications.

To address this inefficiency, recent works have proposed one-step diffusion approaches that distill the multi-step denoising trajectory into a single forward pass~\cite{yin2024improved,yin2024one,lin2025diffusion}. Originally introduced in the context of image generation, these models preserve much of the fidelity and realism of full diffusion while drastically reducing inference latency. More importantly, one-step diffusion has proven particularly effective in restoration-oriented tasks such as image super-resolution~\cite{wu2024one,dong2025tsd}, image deblurring~\cite{liu2025one}, \etc, where the goal is to transform degraded inputs into high-quality outputs. 
Compared with multi-step diffusion, one-step diffusion offers inference efficiency that aligns well with the demands of event-to-video reconstruction, which necessitates generating high-frame-rate videos with both accuracy and speed.

\section{EvDiff}
\label{sec:method}

Before detailing our \ourmodel for event-based video reconstruction (§~\ref{subsec:architecture}), we first formulate the problem and discuss the motivation that drives our design (§~\ref{subsec:motivation}).

\subsection{Motivation}
\label{subsec:motivation}

Instead of capturing full frames at a fixed frame rate, event cameras only asynchronously record per-pixel brightness changes. For each pixel, an event $e = (x,y,t,p)$ is triggered when the intensity ($\mathcal{I}$) variation in log domain exceeds a preset contrast threshold $C$:
\begin{equation}
    \left| \log \mathcal{I}(x,y,t) - \log \mathcal{I}(x,y,t-\delta t) \right| \ge C,
    \label{eq:event_generation}
\end{equation}
where $x$, $y$, $t$, and $p\in\{+1,-1\}$ denote the coordinate, timestamp, and polarity, respectively. This unique mechanism yields high temporal resolution and dynamic range but produces sparse, non-intensity data, \ie, ``events''.

Following the seminal event-based video reconstruction work E2VID~\cite{rebecq2019events,rebecq2019high}, existing SOTA methods typically adopt an end-to-end training paradigm, where a single model $f(\cdot)$ directly maps a sequence of events to a sequence of images with a UNet-like model. 
Although this pipeline delivers substantial advances over earlier optimization-based methods~\cite{munda2018real,scheerlinck2018continuous}, it now encounters a critical methodological bottleneck, as outlined in the aforementioned \myhyperlink{P1}{\blackcircled{1}}
 and \myhyperlink{P2}{\blackcircled{2}}. 

As a result, E2VID-style models exhibit limited ability to recover fine spatial structures and realistic textures, leading to blur, ghosting, and artifacts.

By looking deeper into the ``degradation pattern'' of these results, we observe that E2VID-style models tend to produce relatively stable artifacts: characteristic combinations of blur, edge disruption, and block-like distortions.
%
%
This stems from the inherent initial condition ambiguities.
Therefore, if we can construct a degradation space that simulates these artifacts on images and leverage a powerful diffusion prior for reconstruction, it would bridge the event-to-video conversion task with large-scale datasets and enable large-parameter model training.

Motivated by this, we reformulate the task as
\begin{equation}
    \{\mathbf{I}_i'\}_{i=1}^{N} = f_1(\{e_{\Delta t}\}; \Theta_{1}), \quad \{\hat{\mathbf{I}}_i\}_{i=1}^{N} = f_2(\{\mathbf{I}_i'\}; \Theta_{2}),
    \label{eq:architecture}
\end{equation}
where $\{\mathbf{I}_i'\}_{i=1}^{N}$ are the degraded frames reconstructed from events, and $f_2(\cdot; \Theta_2)$ is a high-capacity generative model pretrained on large-scale image data. This decomposition provides two key advantages: it 1) removes the reliance on scarce paired event–image datasets, and 2) enables the use of large diffusion models with visual priors to achieve high-quality and perceptually consistent reconstructions.

This reformulation establishes the foundation of our framework and training strategy, described in \S\ref{subsec:architecture} and \S~\ref{subsec:distillation}, respectively.

\begin{figure*}[t]  
\centering
\includegraphics[width=\textwidth]{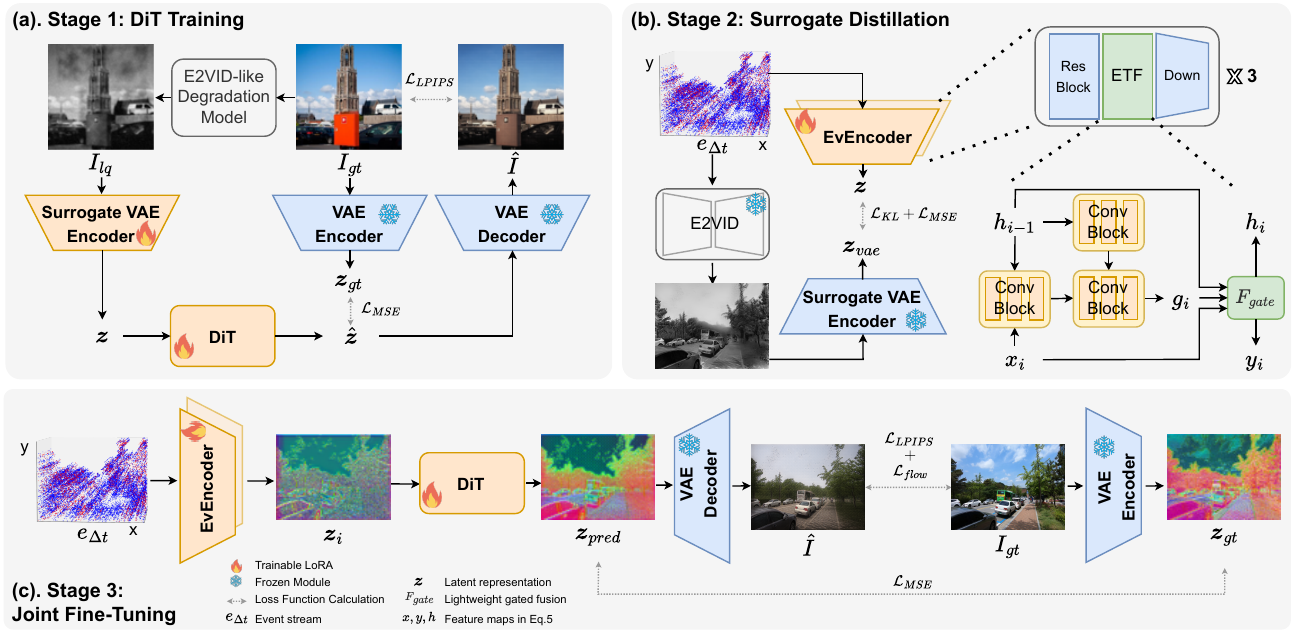}
\caption{\textbf{The proposed Surrogate Training Pipeline}. Stage 1: We train a DiT model and Surrogate VAE Encoder with LQ-HQ pairs; Stage 2: The Surrogate VAE Encoder is distilled into EvEncoder; Stage 3: We finetune the whole EvDiff model.}
\label{fig:pipeline}
\end{figure*}

\subsection{General Architecture of EvDiff}
\label{subsec:architecture}
Following the generic form from Eq.~\ref{eq:architecture}, we propose EvDiff to bring the powerful diffusion prior for high-quality event-based video reconstruction.
EvDiff consists mainly of two parts: a \ourencoder $\encoder$ and a one-step diffusion model OSDiff:
\begin{equation}
    \{\boldsymbol{z}_{i}\}_{i=1}^{N} = \encoder (\{e_{\Delta t}\}), 
    \label{eq:stage_1}
\end{equation}
\begin{equation}
    \{\mathbf{I}_i\}_{i=1}^{N} = \mathcal{D}(\mathrm{OSDiff}(\{\boldsymbol{z}_{vid}\})).
    \label{eq:stage_2}
\end{equation}
where $\mathcal{D}$ is the VAE decoder, and $\{\boldsymbol{z}_{i}\}_{i=1}^{N}$ represents the latent representation.
The training strategy will be detailed in \S\ref{subsec:distillation}.
During inference, however, EvDiff operates in a single-stage, end-to-end manner, as shown in Fig.~\ref{fig:first_page_fig}.

\noindent\textbf{\ourencoder.}
Given an input event stream $\{ e_{\Delta t} \}$, where $\{ e_{\Delta t} \} = \{ e_i \mid t_i \in [t, t+\Delta t) \}$ denotes all events occurring within the temporal window $\Delta t$, we first encode them into latent representation. 
To be specific, asynchronous event streams are first converted to synchronous voxel grids~\cite{zhu2018ev}
$\textbf{V} \in \mathbb{R}^{H \times W \times T}$
which contain temporal information in $T$ channel. 
The resulting voxels are encoded by the \ourencoder to obtain latent representation for further processing. 
The most intuitive idea to reconstruct the videos with $T$ frames with a diffusion model is to send all events into the diffusion model with ControlNet~\cite{zhang2023adding} for self-attention, similar to SVD~\cite{blattmann2023stable}, with a computational complexity of $\mathcal{O}\left((T H W)^2 \cdot C\right)$. However, one of the most significant advantages of event cameras is the high temporal resolution. Thus, event-based video reconstruction methods typically focus on generating high-speed and high-frame-rate videos that conventional cameras cannot capture~\cite{rebecq2019high}, which substantially increases the temporal length $T$ and consequently leads to unacceptable computational overhead.

Therefore, we design a recurrent architecture~\cite{hochreiter1997long,lstm} in \ourencoder to efficiently keep the temporal information flow.
For efficiency, we propose the Efficient Temporal Fusion (ETF) module in \ourencoder, which employs a lightweight gated fusion mechanism~\cite{lstm}, as shown in Fig.~\ref{fig:pipeline} (b). Each ETF module contains three convolutional blocks to extract features from both historical and current data, generating a gate weight that adaptively balances their contributions:
\begin{equation}
    \mathbf{y}_{i} = \mathbf{g}_{\text{i}} \odot \mathbf{x}_{\text{i}} + (1 - \mathbf{g}_{\text{i}}) \odot \mathbf{h}_{\text{i-1}},
    \label{eq:rnn}
\end{equation}
where $\mathbf{x}_{\text{i}}$ and $\mathbf{h}_{\text{i-1}}$ denote the current and previous feature maps, and $\mathbf{g}_{\text{i}}$ represents the learned fusion gate controlling temporal blending. With the ETF module, we reduce the recomputation from $\mathcal{O}\left((T H W)^2 \cdot C\right)$ to $\mathcal{O}\left((H W)^2 \cdot T \cdot C\right)$ and strike a balance between temporal information flow and computational cost.

\begin{figure*}[t] 
\centering
\includegraphics[width=\textwidth]{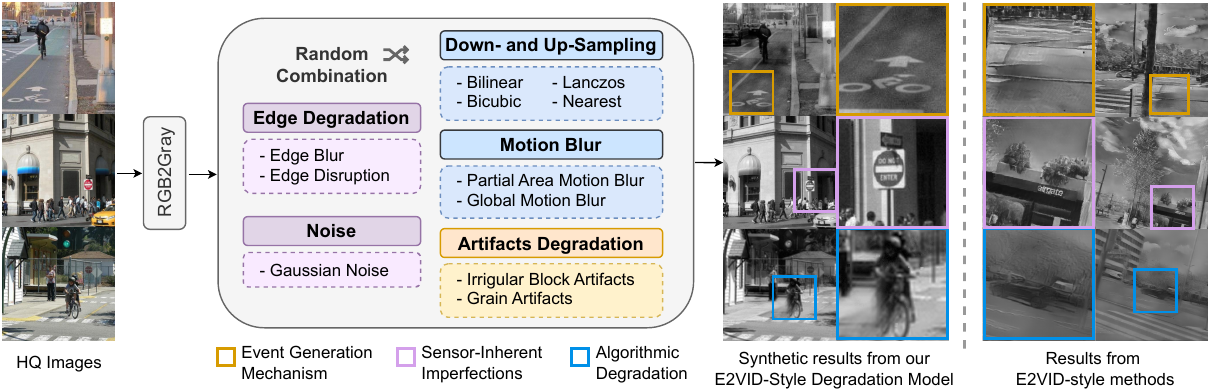}
\caption{\textbf{Overview of the proposed E2VID-Style Degradation Model.} 
HQ images sampled from the Places365~\cite{places365} dataset are cropped to $512 \times 512$ and then processed by our degradation model to generate corresponding LQ images. For comparison, E2VID-style results from the BS-ERGB dataset are shown on the right.
}
\label{fig:degradation}
\end{figure*}

\noindent\textbf{One-Step Diffusion Model.}
Existing SD-based methods usually take random Gaussian noise as the starting point and require multiple diffusion steps. Like aforementioned reasons, this expensive computational cost is not compatible with event cameras. Recent works~\cite{wu2024one,dong2025tsd} prove that a low-quality image can be taken as a starting point for one-step diffusion. Inspired by this, we utilize Stable Diffusion 3 (SD3)~\cite{esser2024scaling} with powerful natural image priors as foundation model. By aligning $\boldsymbol{z}_{i}$ with latent representation from E2VID-style degradated images (see \S\ref{subsec:distillation}), we conduct one-step diffusion for reconstruction:
\begin{equation}
    \hat{\boldsymbol{z}}_i
    = \mathrm{OSDiff}(\boldsymbol{z}_i)
    \triangleq
    \frac{
        \boldsymbol{z}_i - 
        \beta_{t^*}\,
        \boldsymbol{\epsilon}(
            \boldsymbol{z}_i;
            t^*
        )
    }{
        \alpha_{t^*}
    },
    \label{equ:osdiff}
\end{equation}
where $\boldsymbol{\epsilon}$ denotes a denoising network, $\alpha_{t^*}$ and $\beta_{t^*}$ are scalar coefficients determined by the fixed diffusion timestep $t^*$. We simplify $\boldsymbol{z}_{i}$ as $\boldsymbol{z}$ as we discuss only single frame case in the rest part.
Then, we have the final prediction by:
\begin{equation}
    \hat{\mathbf{I}} = \mathcal{D}(\hat{\boldsymbol{z}}).
    \label{eq:decoder}
\end{equation}

So far, we have split the typical E2VID-style paradigm into two parts and introduced large diffusion models into the event-based video reconstruction task, which answers \myhyperlink{P1}{\blackcircled{1}}. In the following sections, we describe how to address the scarcity of event–image paired data when training EvDiff.

\subsection{Surrogate Training Pipeline}
\label{subsec:distillation}
As illustrated in \S\ref{subsec:motivation}, there is a highly fixed degradation pattern in results from regression-based event-to-video converting methods. Hence, we use synthetic E2VID-style degradation images as a ``surrogate'' and propose our Surrogate Training Pipeline that enables large-scale data training, as shown in Fig.~\ref{fig:pipeline}.

\paragraph{Stage 1: DiT Training.}
Given large-scale image dataset, we first synthesize the E2VID-style degraded images with the degradation model in \S\ref{subsec:data_synthesize}. The paired images are feed into a Diffusion Transformer (DiT) for one-step diffusion training. Following the practices of diffusion-based image restoration methods~\cite{chen2024faithdiff,yu2024scaling}, we also make the VAE encoder (Surrogate VAE encoder) trainable to better align the latent representations with the distribution of degraded inputs. 
The training objective combines latent reconstruction and perceptual consistency:
\begin{equation}
    \mathcal{L}_{\text{DiT}} = \lambda_1 \|\boldsymbol{z}_{\text{pred}} - \boldsymbol{z}_{\text{gt}}\|_2^2 + \lambda_2 \text{LPIPS}(\mathbf{\hat{I}}, \mathbf{I}_{\text{gt}}).
    \label{eq:dit_loss}
\end{equation}

\paragraph{Stage 2: Surrogate Distillation.}
We then replace the current Surrogate VAE encoder with our \ourencoder, which is specifically designed to adapt to event data. As shown in Fig.~\ref{fig:pipeline}(b), during the distillation, the event data is first converted to E2VID-style degradation images with an off-the-shelf E2VID model, and consequently encoded to $\boldsymbol{z}_{vae}$. As student model, \ourencoder encodes event data to $\boldsymbol{z}$ directly. Training follows the distillation loss:

\begin{equation}
    \mathcal{L}_{\text{distill}} = \text{KL}(q_{\phi}(\boldsymbol{z}|E)) + \|\boldsymbol{z} - \boldsymbol{z}_{\text{vae}}\|_2^2.
    \label{eq:distillation_loss}
\end{equation}


\definecolor{tableRowGray}{gray}{0.97}

\begin{table*}[!t]
\centering
\small
\setlength{\tabcolsep}{10pt}
\renewcommand\arraystretch{1.1}
\caption{\textbf{Quantitative comparison} on BS-ERGB~\cite{tulyakov2022time} and DSEC~\cite{dsec}, with the best results highlighted in \best{red} and the second best in \second{blue}.}

\resizebox{\textwidth}{!}{%
    \begin{tabular}{r||ccccc||ccccc}
    \bottomrule[0.12em]
    \rowcolor{tableHeadGray}
    \textbf{Method} & \textbf{MSE$\downarrow$} & \textbf{SSIM$\uparrow$} & \textbf{LPIPS$\downarrow$} & \textbf{FID$\downarrow$} & \textbf{FVD$\downarrow$} & \textbf{MSE$\downarrow$} & \textbf{SSIM$\uparrow$} & \textbf{LPIPS$\downarrow$} & \textbf{FID$\downarrow$} & \textbf{FVD$\downarrow$} \\ \hline \hline
    & \multicolumn{5}{c||}{\textbf{BS-ERGB}} & \multicolumn{5}{c}{\textbf{DSEC}} \\ \hline 

    \rowcolor{tableRowGray}
    E2VID\pub{CVPR2019} & 0.1175 & 0.3313 & 0.5518 & 276 & 1688 & 0.1042 & 0.3393 & 0.5427 & 239 & 1541 \\

    FireNet\pub{WACV2020} & 0.0857 & 0.3302 & 0.5328 & 271 & 1909 & 0.0927 & 0.3478 & 0.5670 & 262 & 1492 \\

    \rowcolor{tableRowGray}
    E2VID+\pub{ECCV2020} & 0.0717 & \best{0.3710} & \second{0.4397} & \second{238} & 1652 & 0.0881 & 0.2812 & 0.5138 & 248 & \second{1397} \\

    FireNet+\pub{ECCV2020} & 0.0827 & 0.3080 & 0.4903 & 313 & \second{1562} & 0.0913 & 0.2268 & 0.5655 & 291 & 1592 \\

    \rowcolor{tableRowGray}
    SPADE-E2VID\pub{TIP2021} & 0.0891 & 0.3235 & 0.6715 & 347 & 2343 & \second{0.0601} & \best{0.4578} & \second{0.4911} & \second{234} & \best{1277} \\

    SSL-E2VID\pub{CVPR2021} & 0.0798 & 0.3448 & 0.6169 & 312 & 1907 & 0.1055 & 0.3303 & 0.5565 & 326 & 1521 \\

    \rowcolor{tableRowGray}
    ET-Net\pub{ICCV2021} & \second{0.0683} & \second{0.3557} & 0.4566 & 270 & 1569 & 0.0737 & 0.2953 & 0.5232 & 259 & 1496 \\

    HyperE2VID\pub{TIP2024} & 0.0756 & 0.3489 & 0.4598 & 275 & 1673 & 0.0691 & 0.2977 & 0.5297 & 272 & 1445 \\

    \rowcolor{tableRowGray}
    \textbf{EvDiff~(Ours)} & \best{0.0463} & 0.3394 & \best{0.4023} & \best{148} & \best{984} & \best{0.0476} & \second{0.3677} & \best{0.4226} & \best{129} & 1491 \\

    \hline
    \end{tabular}}
\label{tab:results}
\end{table*}

\paragraph{Stage 3: Joint Fine-Tuning.}
With two main parts ready, we jointly fine-tune the whole model using small-scale event–RGB paired dataset. 
This stage adapts the model to genuine event distributions and enforces temporal coherence under realistic motion patterns. The joint loss integrates perceptual, flow-based temporal, and latent consistency terms:
\begin{equation}
    \mathcal{L}_{\text{joint}} = \lambda_1 \mathcal{L}_{\text{LPIPS}} + \lambda_2 \mathcal{L}_{\text{flow}} + \lambda_3 \|\boldsymbol{z}_{\text{pred}} - \boldsymbol{z}_{\text{gt}}\|_2^2.
    \label{eq:joint_loss}
\end{equation}

Although our training strategy follows a multi-stage paradigm, during inference, EvDiff predicts the reconstruction results from the event stream in a single step.
Our sophisticated designed Surrogate Training Pipeline shifts the need for large-scale event–image paired data (unavailable) to widely accessible large-scale image datasets.
This design enables the training of large diffusion-based foundation models for event reconstruction.
The key lies in the notion of the ``surrogate'' — the E2VID-style degraded images, which we will introduce next.

\subsection{E2VID-Style Degradation Model}
\label{subsec:data_synthesize}
We use the term “E2VID-style” to denote the common degradation characteristics observed across modern event video reconstruction models (EVRMs), including E2VID, ETNet, HyperE2VID, and related methods. It does not refer exclusively to the original E2VID architecture. Since the results from E2VID-style regression-based methods demonstrate similar degradation patterns (Fig.~\ref{fig:degradation} (b)), by analyzing the factors leading to these degradations, we can synthesize low-quality (LQ) images with E2VID-style degradation from high-quality (HQ) images.

Through the whole pipeline from event to the final images, We identify three primary degradation factors in the designed E2VID-style model: event generation mechanism, sensor-inherent imperfections, and algorithmic degradation.

\noindent\textbf{Event Generation Mechanism(EGM).}
Indicated by the first princinple of event camera (Eq.~\ref{eq:event_generation}), events only contain brightness change information, without initial condition, or absolute intensity values. This ambiguity leads to blotchy and uneven textures. To synthesize it, we first identify low-detail regions according to local variance and gradient magnitude, selecting smooth areas via percentile-based thresholding. Within these regions, irregular dark patches are stochastically generated by blending ellipses and polygons with layered noise, while multi-scale granularity enhances perceptual roughness and local contrast.

\noindent\textbf{Sensor-Inherent Imperfections(SII).}
Due to the inherent finite pixel temporal bandwidth, trailing events persist after the actual brightness change~\cite{liu2024seeing}. Furthermore, random fluctuations in photon arrival and circuit leakage induce spurious events that degrade the signal-to-noise ratio of the event stream~\cite{whitepaper}. These two lead to edge degradation and noise. We synthesize them by applying soft blurring and stochastic discontinuities to locally displace boundary pixels, and adding Gaussian noise.

\noindent\textbf{Algorithmic Degradation(AD).}
When the motion in the scene is too fast, or the accumulation time interval for events is too long, motion blur is introduced during voxelization. Moreover, event-to-image reconstruction is an under-determined inverse problem with limited spatial bandwidth. The voxel aggregation and the finite-capacity reconstruction network implicitly suppress high-frequency details and introduce spatial uncertainty.

To approximate these effects, we incorporate motion blur and a resize operation (down-sampling followed by up-sampling) into our degradation model. The resize operation simulates the spatial low-pass filtering and resolution bottleneck introduced by voxel accumulation and reconstruction, leading to detail attenuation and structural smoothing similar to those observed in E2VID outputs.

To summarize, given the HQ image, we first convert it to grayscale and apply all degradation factors with random shuffling to model E2VID-style degradation. This enables training on large-scale datasets and addresses \myhyperlink{P2}{\blackcircled{2}}.

Despite its effectiveness, the proposed framework still has several limitations:

\begin{itemize}
    \item \textbf{Residual temporal flicker.} Our framework employs a frame-wise diffusion model and encourages temporal consistency mainly through latent consistency across frames. While this improves stability, slight illumination or appearance flicker may still occur in dynamic scenes.

    \item \textbf{Backbone-induced structural artifacts.} Since our implementation adopts SD3 as the generative backbone, we occasionally observe faint periodic grid-like artifacts in reconstructed frames. This is related to the patch-based architecture of the underlying DiT model and has also been reported in other SD3-based works.
\end{itemize}


\begin{figure}[!t] 
\centering
\setlength{\tabcolsep}{0.5pt} 

\begin{tabular}{cccccc}

\includegraphics[width=0.163\textwidth]{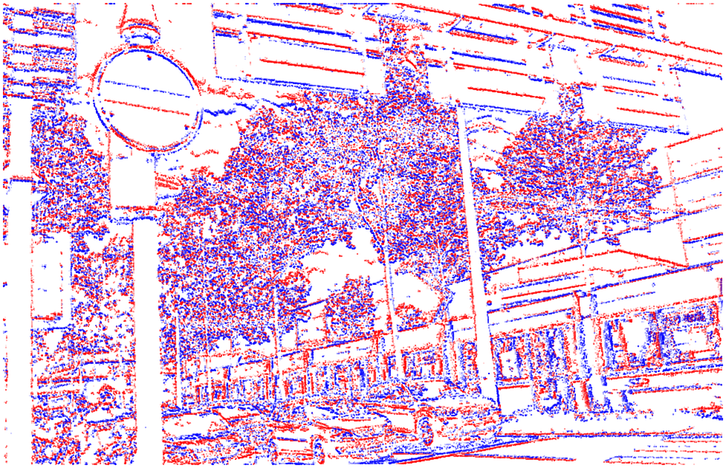} &
\includegraphics[width=0.163\textwidth]{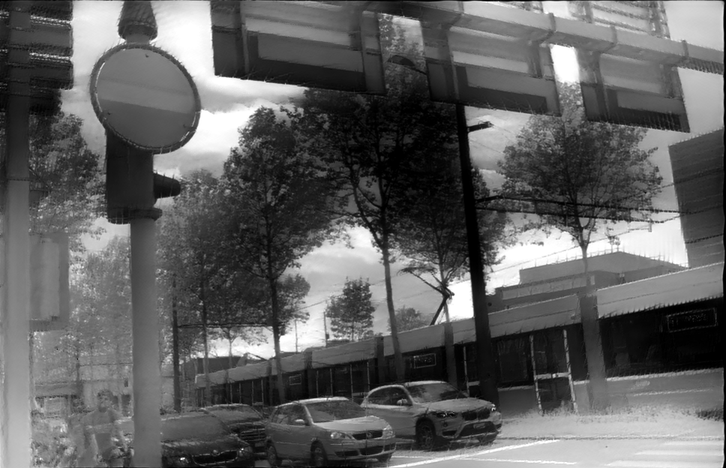} &
\includegraphics[width=0.163\textwidth]{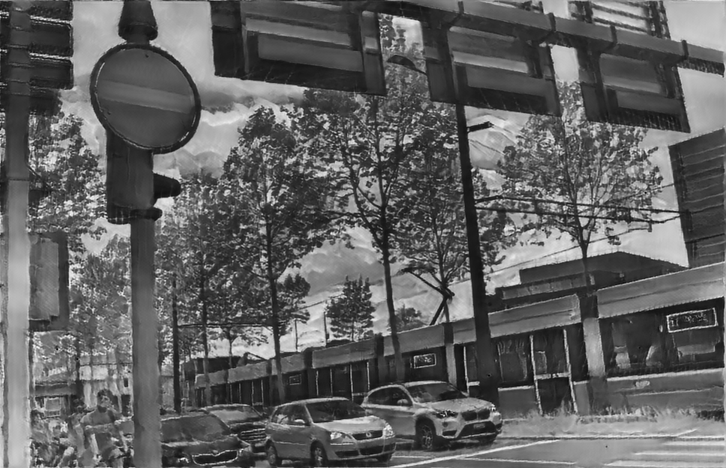} &
\includegraphics[width=0.163\textwidth]{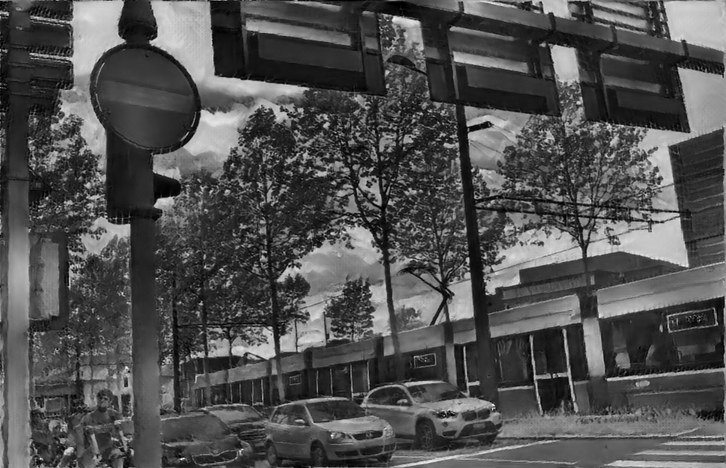} &
\includegraphics[width=0.163\textwidth]{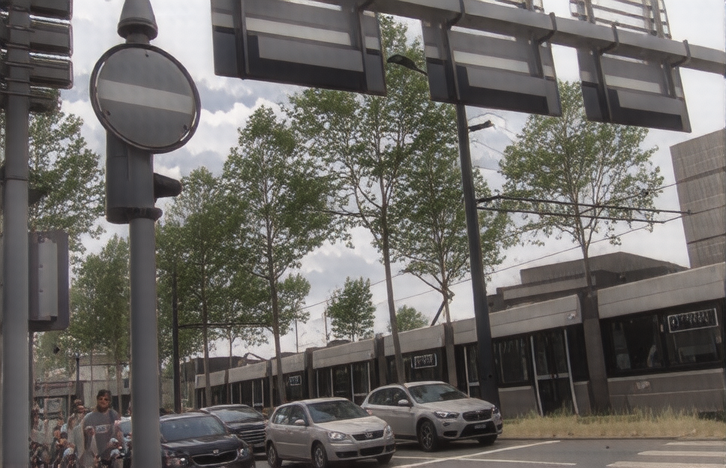} &
\includegraphics[width=0.163\textwidth]{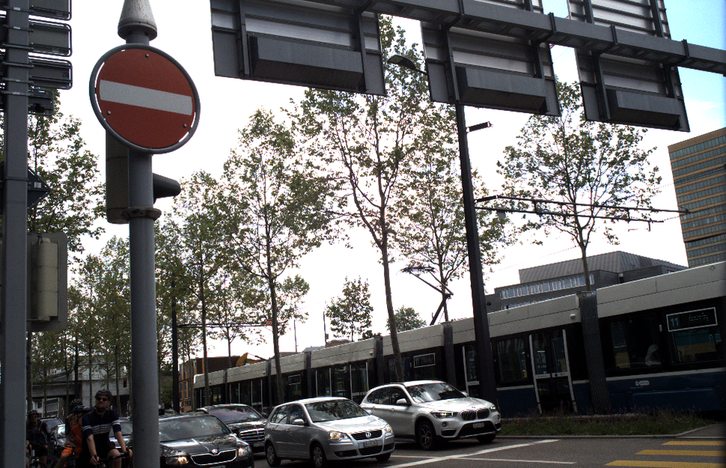} \vspace{-2.5pt} \\ 

\includegraphics[width=0.163\textwidth]{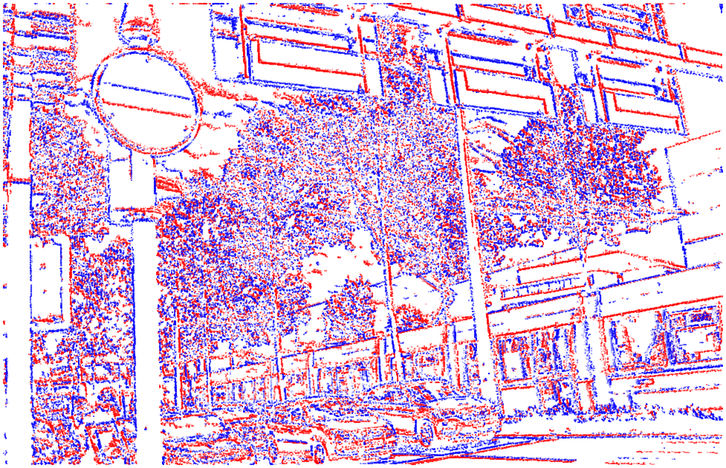} &
\includegraphics[width=0.163\textwidth]{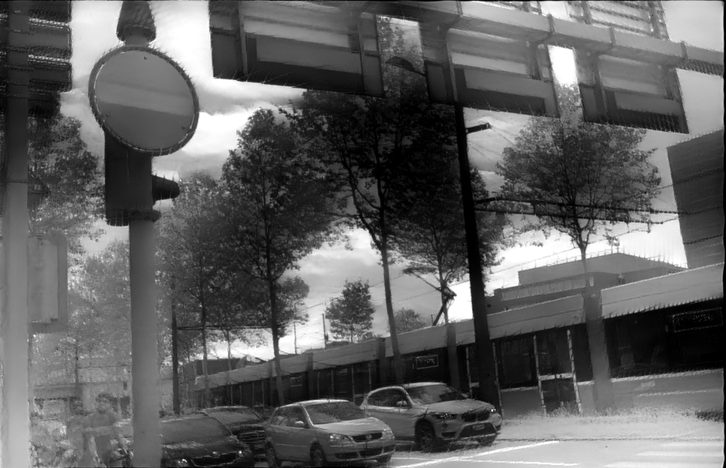} &
\includegraphics[width=0.163\textwidth]{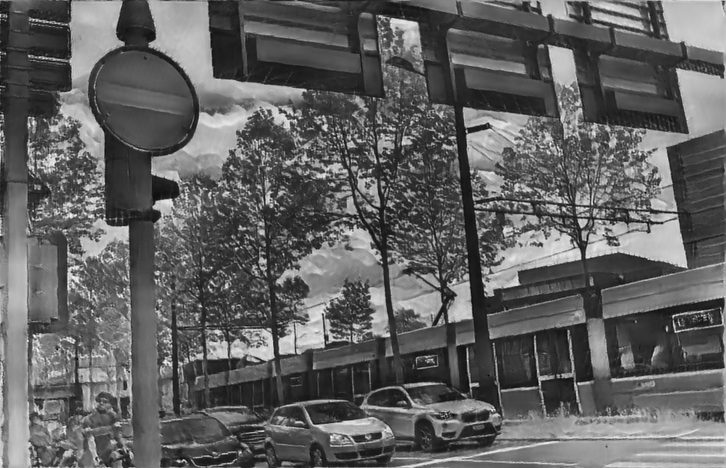} &
\includegraphics[width=0.163\textwidth]{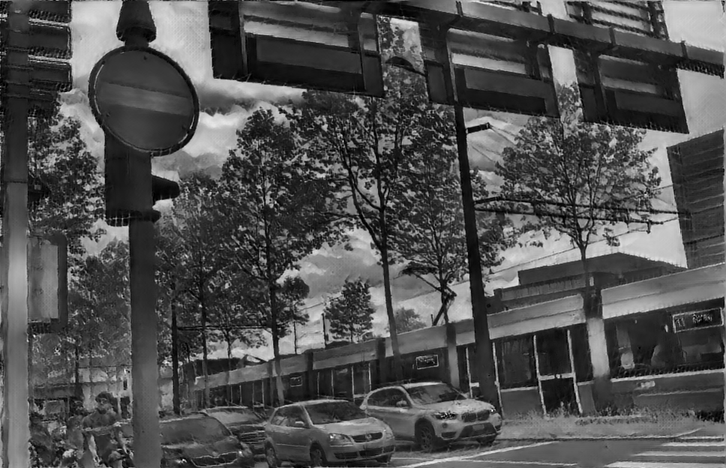} &
\includegraphics[width=0.163\textwidth]{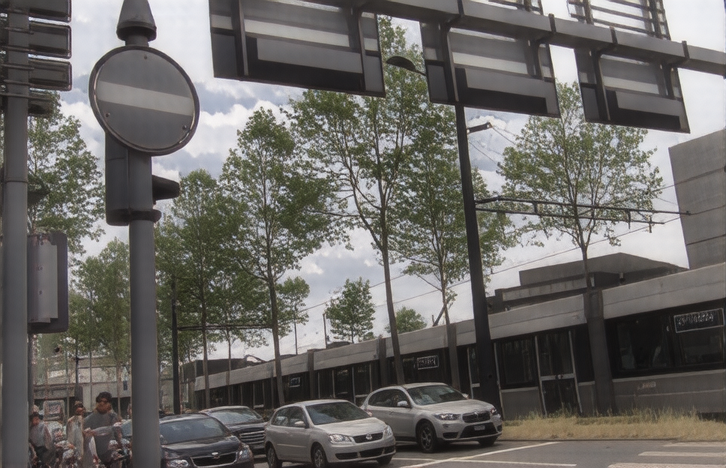} &
\includegraphics[width=0.163\textwidth]{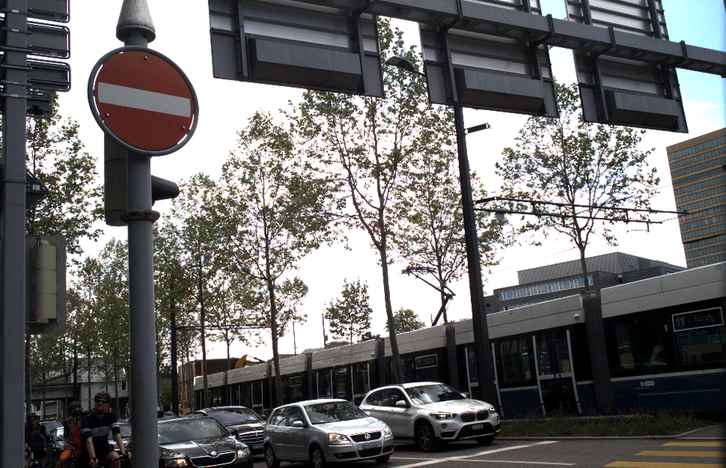} \vspace{-2.5pt} \\ 

\includegraphics[width=0.163\textwidth]{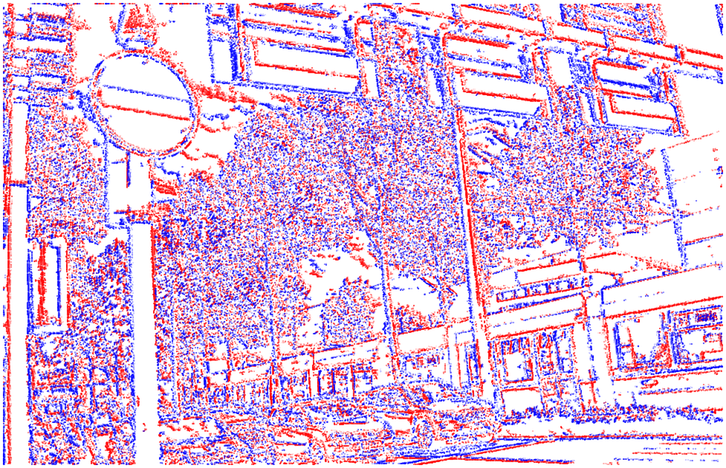} &
\includegraphics[width=0.163\textwidth]{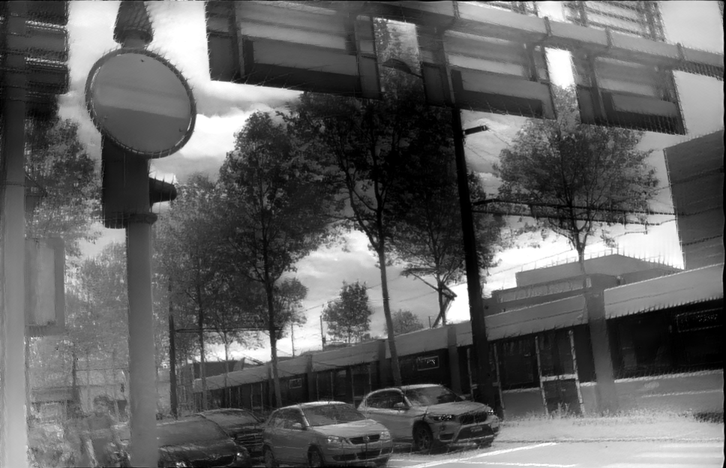} &
\includegraphics[width=0.163\textwidth]{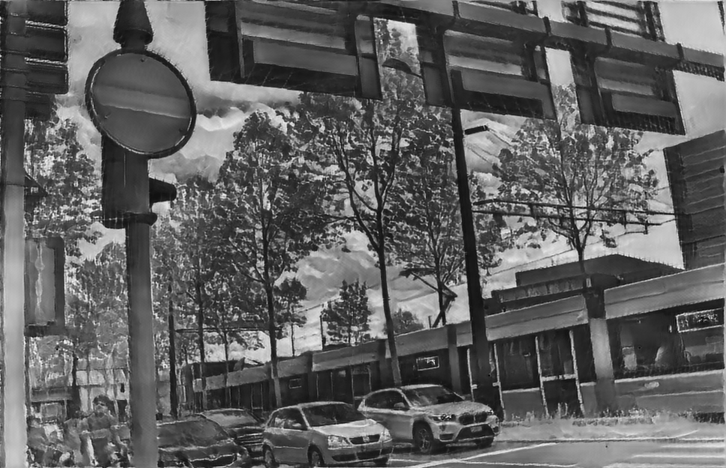} &
\includegraphics[width=0.163\textwidth]{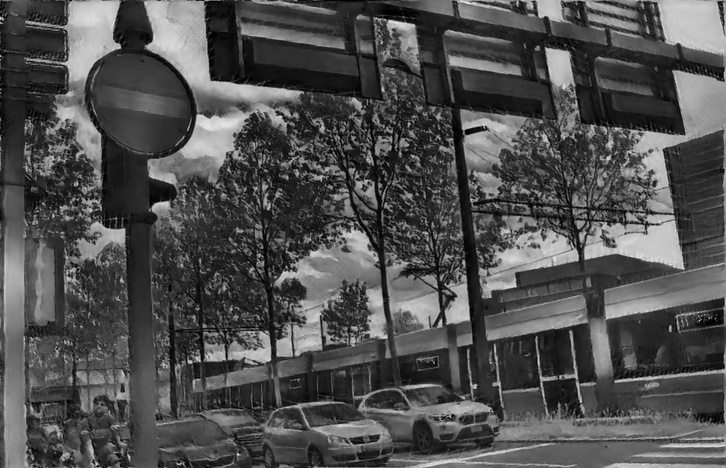} &
\includegraphics[width=0.163\textwidth]{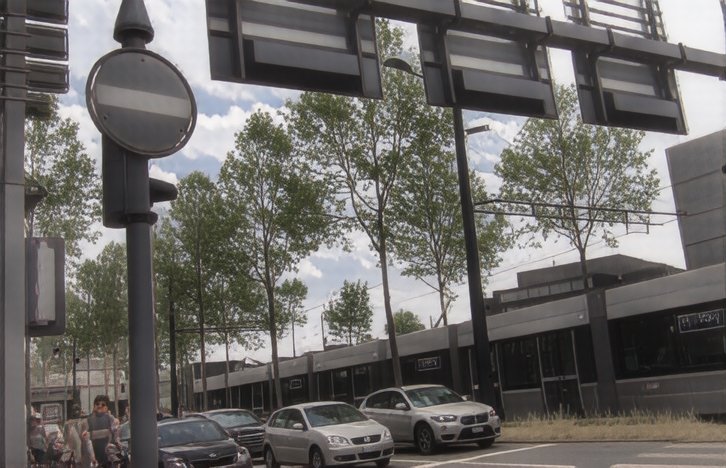} &
\includegraphics[width=0.163\textwidth]{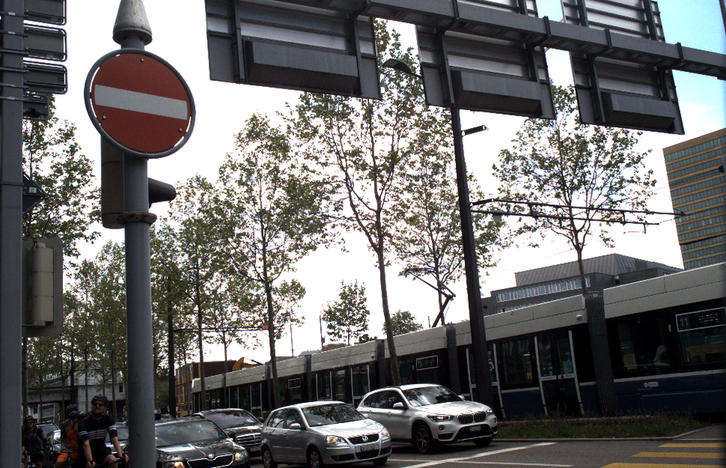} \vspace{-2.5pt} \\ 

\includegraphics[width=0.163\textwidth]{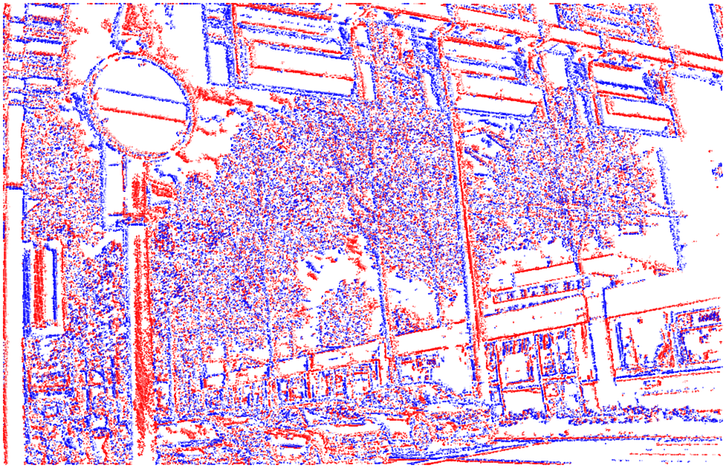} &
\includegraphics[width=0.163\textwidth]{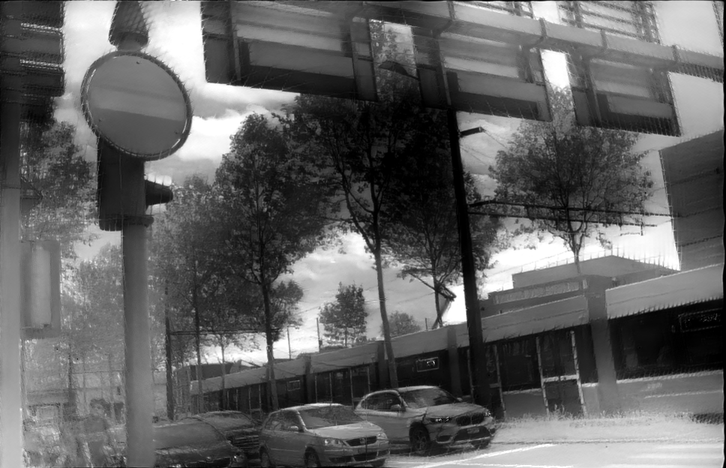} &
\includegraphics[width=0.163\textwidth]{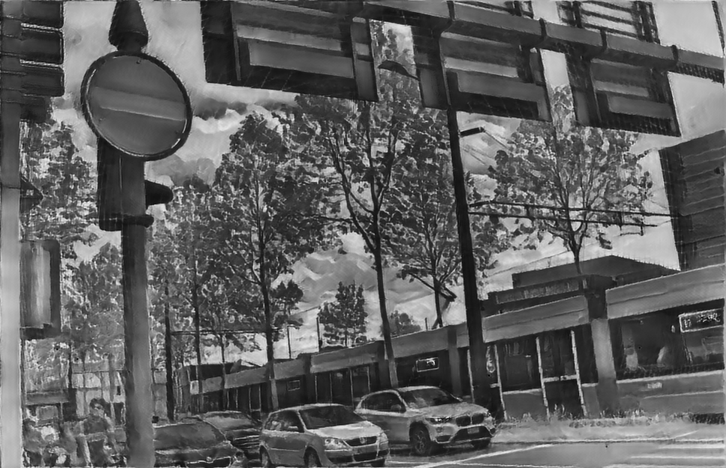} &
\includegraphics[width=0.163\textwidth]{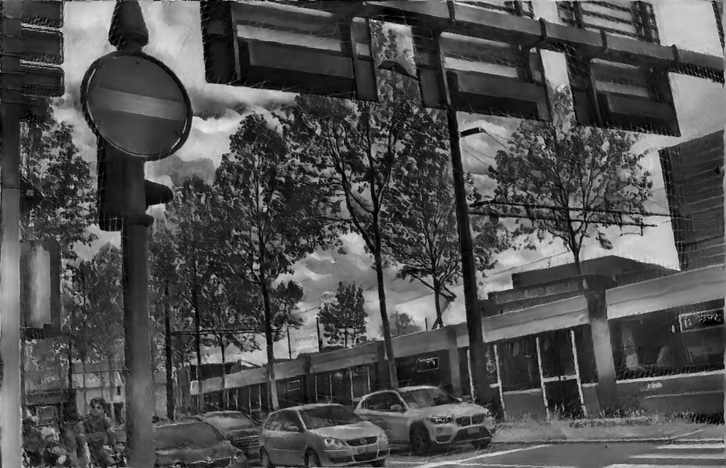} &
\includegraphics[width=0.163\textwidth]{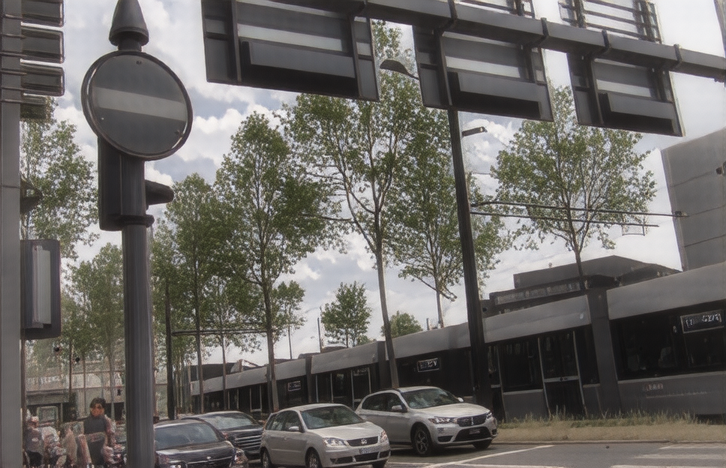} &
\includegraphics[width=0.163\textwidth]{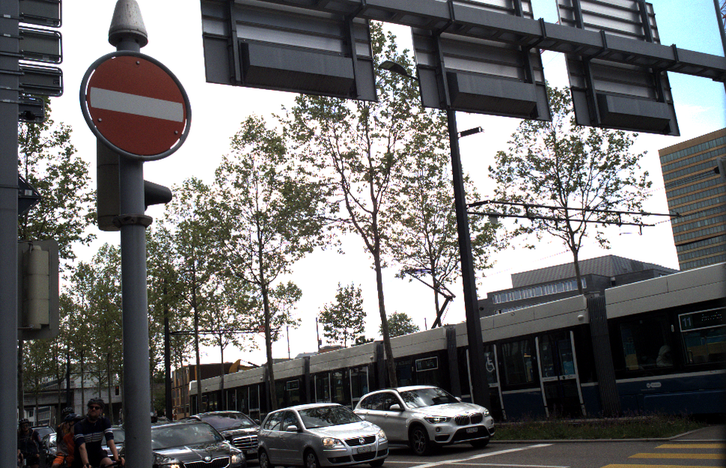} \vspace{-2.5pt} \\ 

\includegraphics[width=0.163\textwidth]{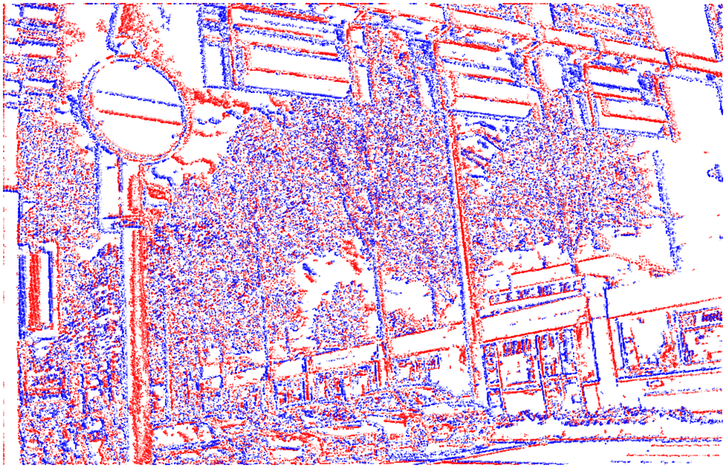} &
\includegraphics[width=0.163\textwidth]{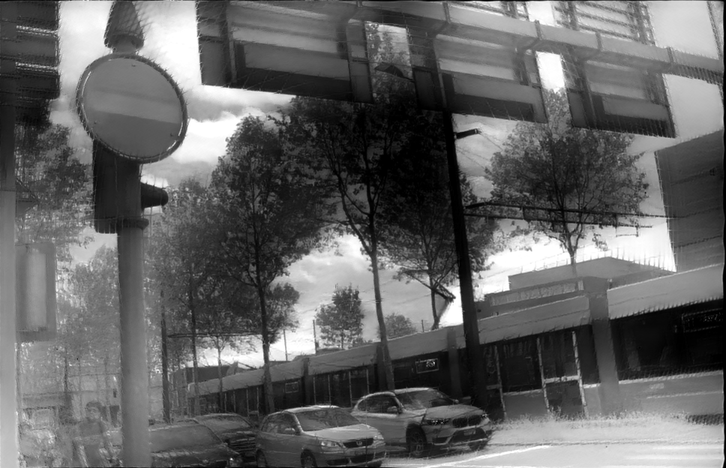} &
\includegraphics[width=0.163\textwidth]{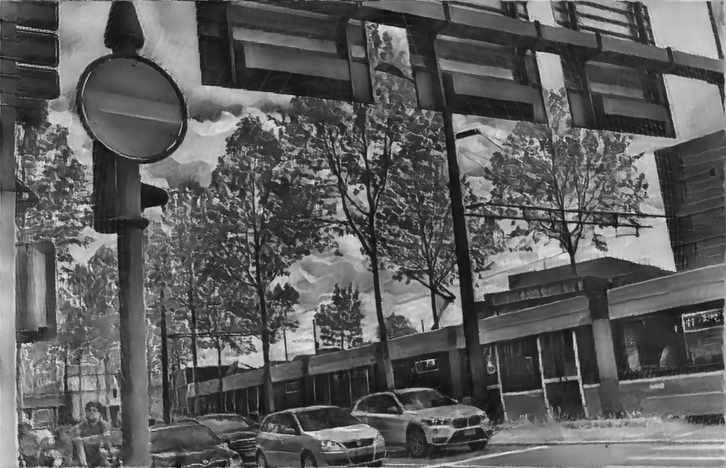} &
\includegraphics[width=0.163\textwidth]{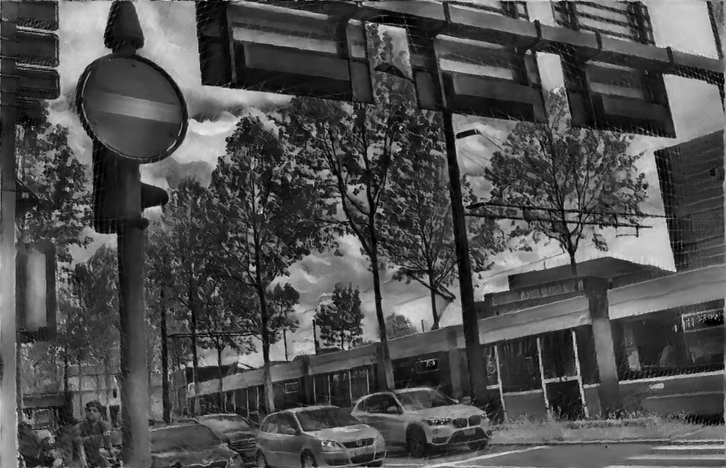} &
\includegraphics[width=0.163\textwidth]{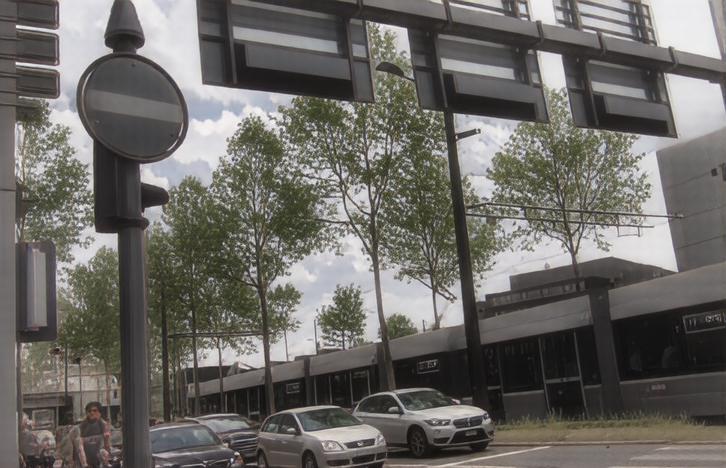} &
\includegraphics[width=0.163\textwidth]{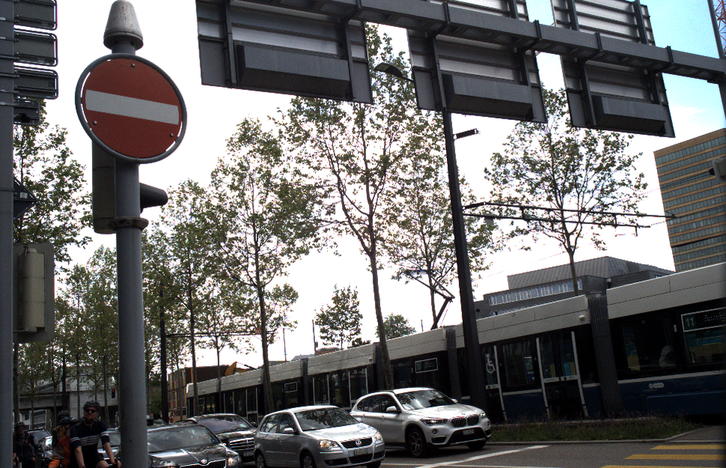} \vspace{-2.5pt} \\ 

\includegraphics[width=0.163\textwidth]{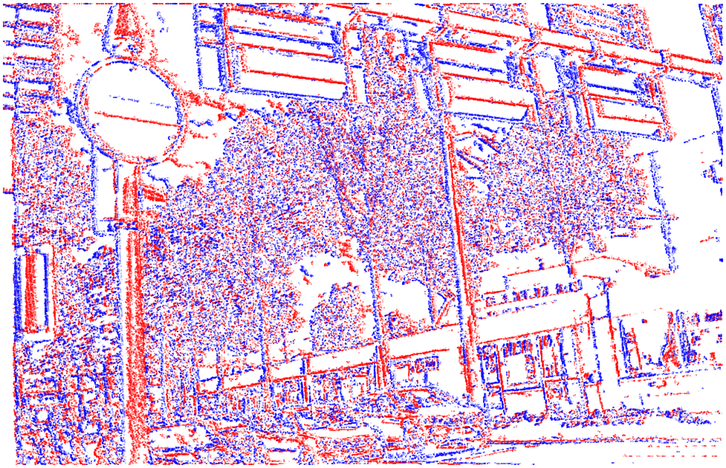} &
\includegraphics[width=0.163\textwidth]{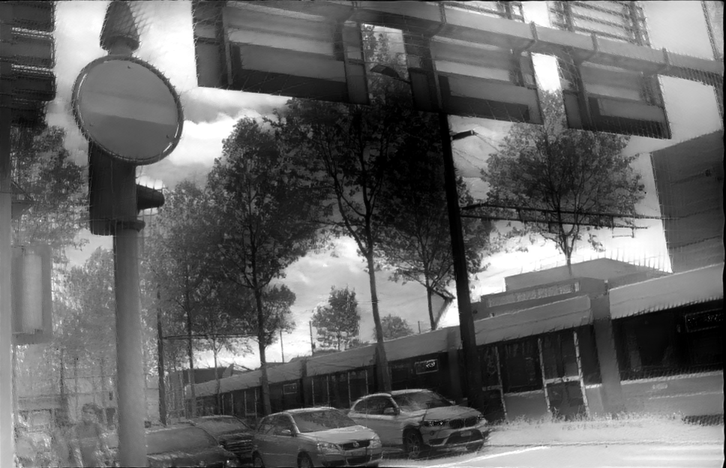} &
\includegraphics[width=0.163\textwidth]{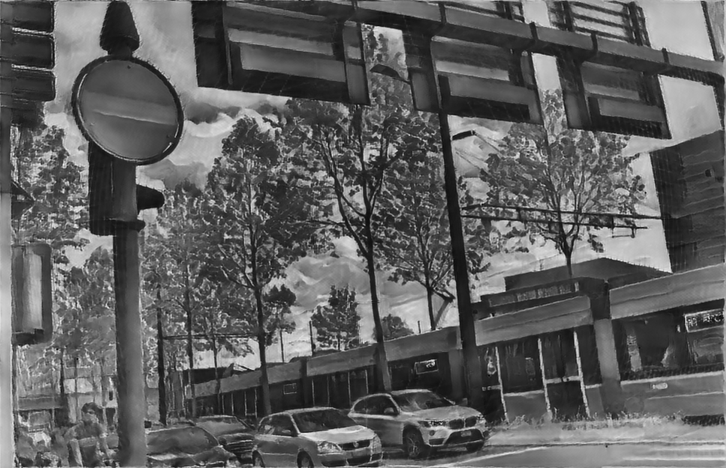} &
\includegraphics[width=0.163\textwidth]{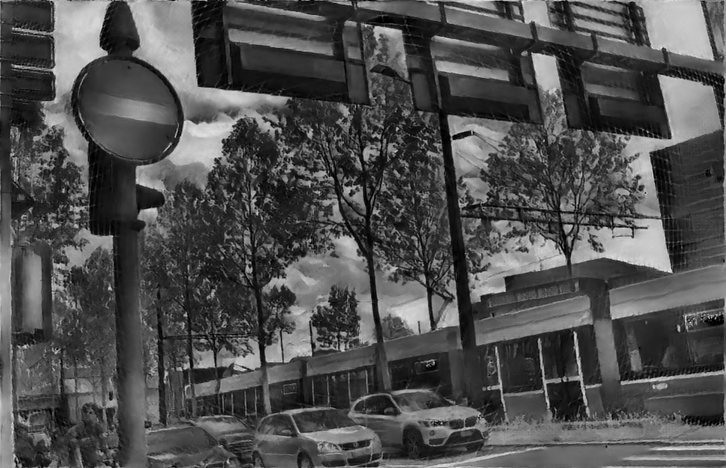} &
\includegraphics[width=0.163\textwidth]{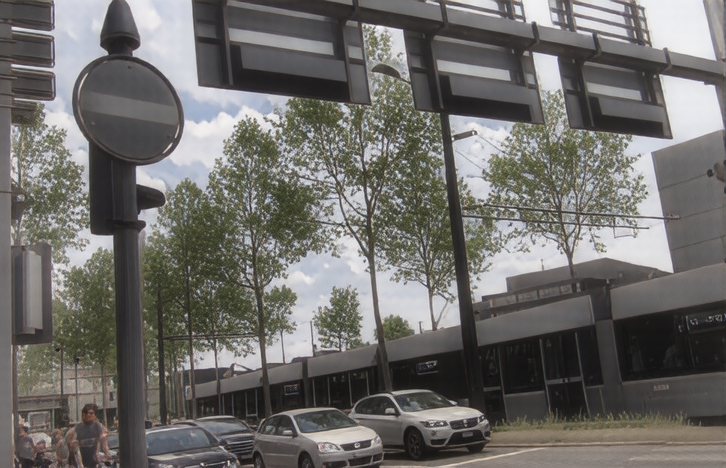} &
\hspace{-5pt}
\includegraphics[width=0.163\textwidth]{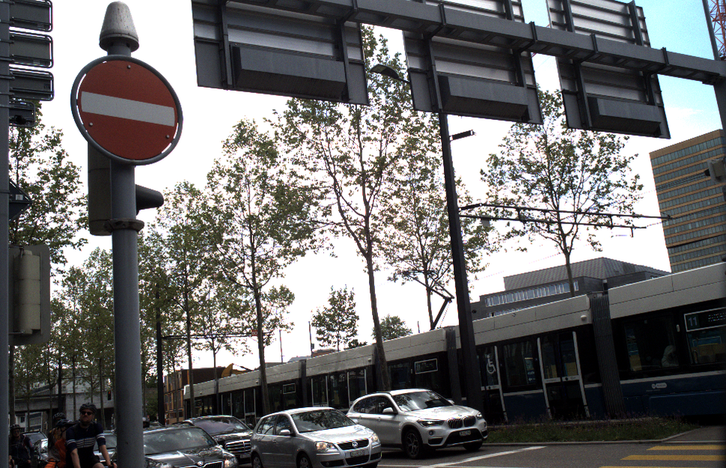} \\

\multicolumn{6}{c}{} \\[-1em]

\includegraphics[width=0.163\textwidth]{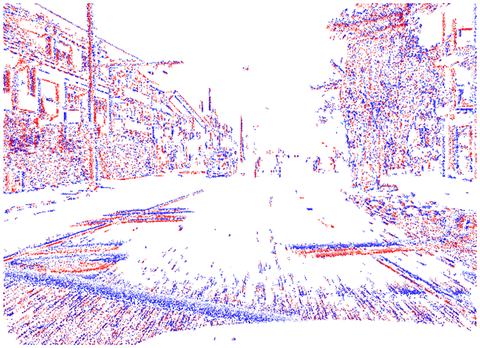} &
\includegraphics[width=0.163\textwidth]{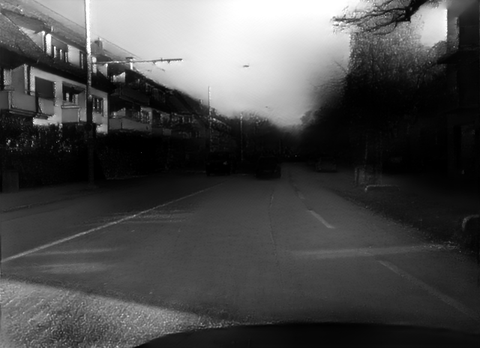} &
\includegraphics[width=0.163\textwidth]{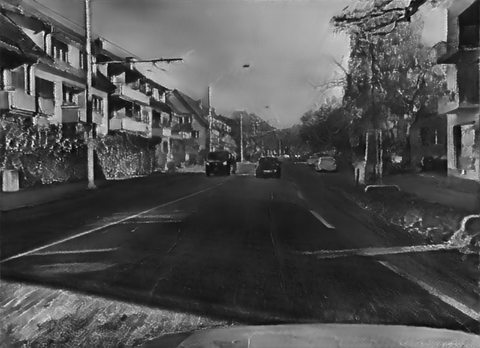} &
\includegraphics[width=0.163\textwidth]{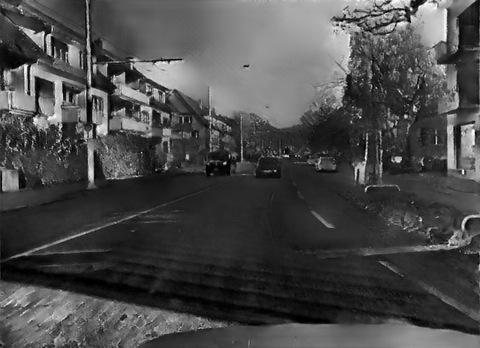} &
\includegraphics[width=0.163\textwidth]{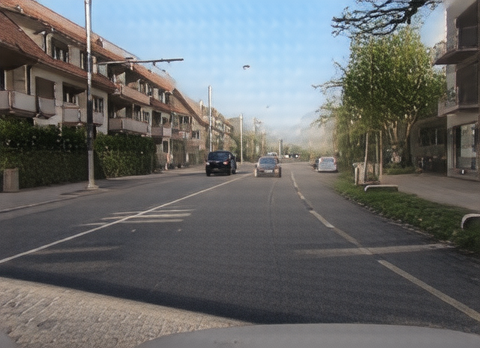} &
\includegraphics[width=0.163\textwidth]{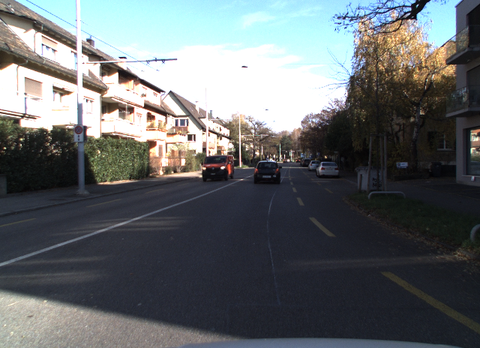} \vspace{-2.5pt} \\ 

\includegraphics[width=0.163\textwidth]{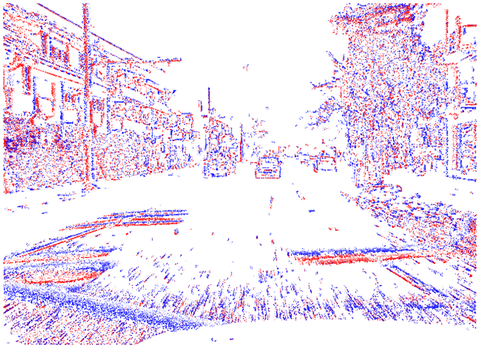} &
\includegraphics[width=0.163\textwidth]{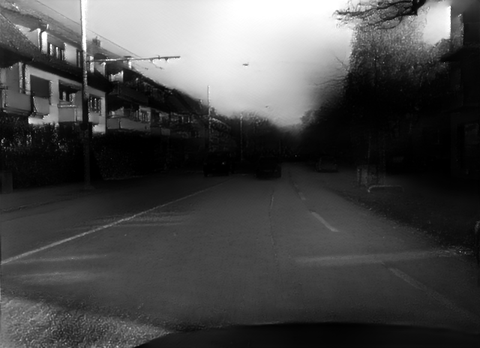} &
\includegraphics[width=0.163\textwidth]{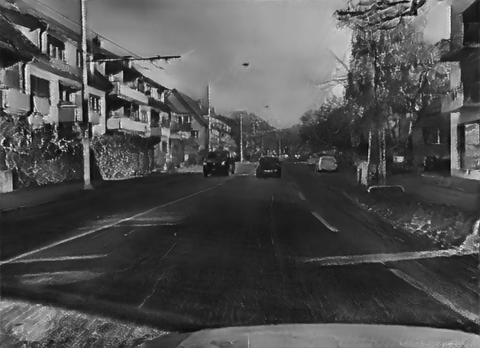} &
\includegraphics[width=0.163\textwidth]{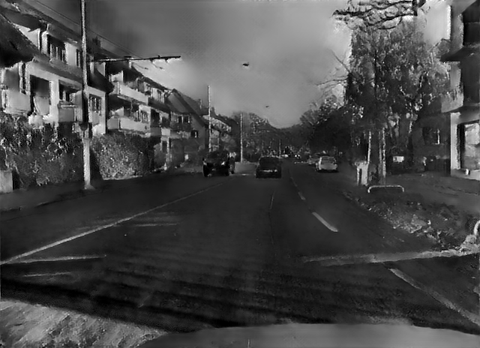} &
\includegraphics[width=0.163\textwidth]{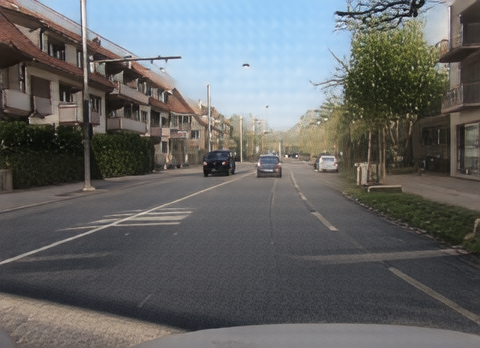} &
\includegraphics[width=0.163\textwidth]{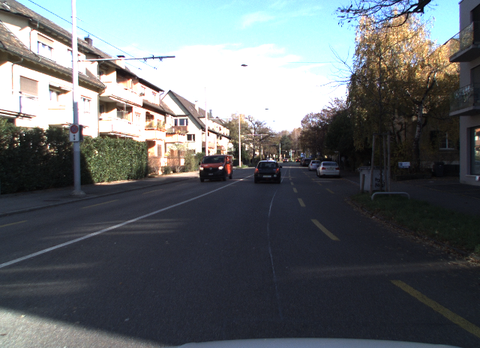} \vspace{-2.5pt} \\ 

\includegraphics[width=0.163\textwidth]{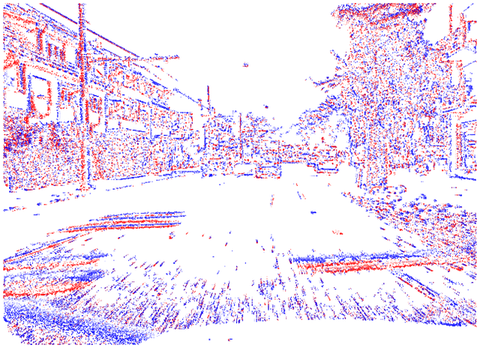} &
\includegraphics[width=0.163\textwidth]{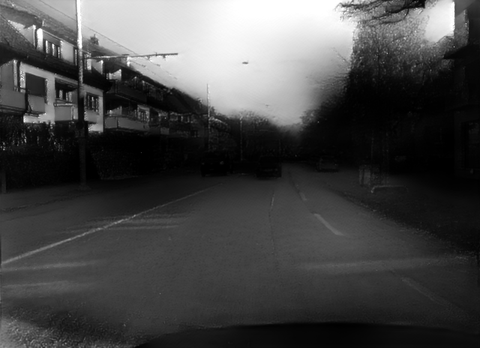} &
\includegraphics[width=0.163\textwidth]{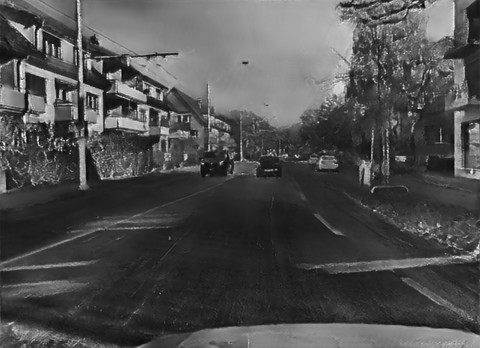} &
\includegraphics[width=0.163\textwidth]{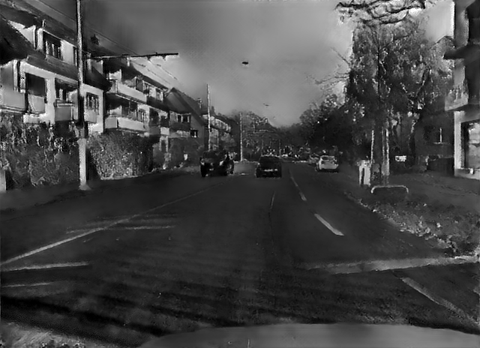} &
\includegraphics[width=0.163\textwidth]{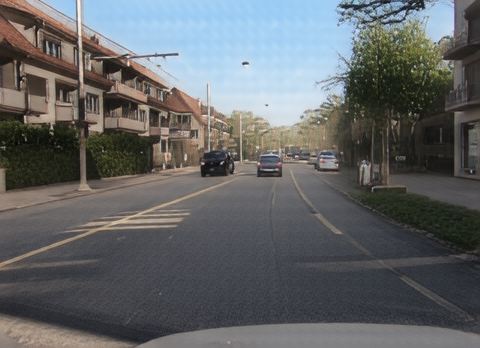} &
\includegraphics[width=0.163\textwidth]{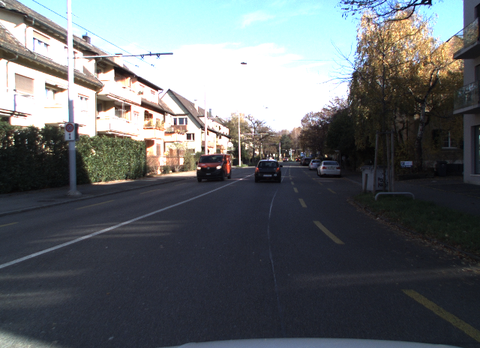} \vspace{-2.5pt} \\ 

\includegraphics[width=0.163\textwidth]{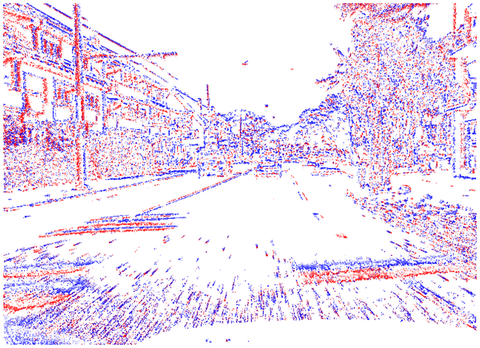} &
\includegraphics[width=0.163\textwidth]{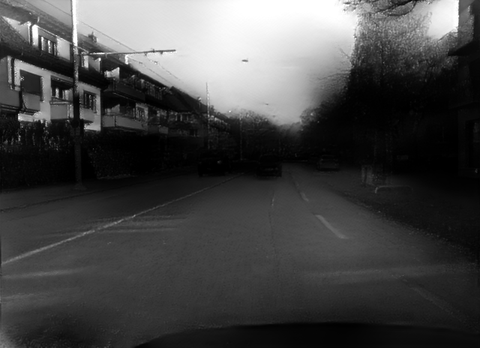} &
\includegraphics[width=0.163\textwidth]{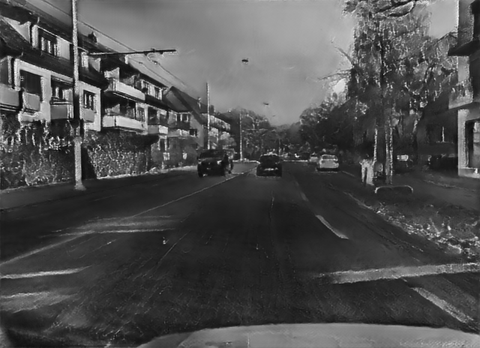} &
\includegraphics[width=0.163\textwidth]{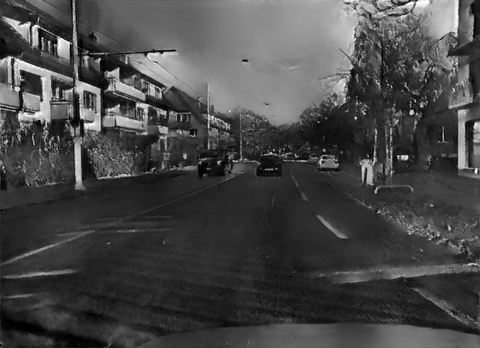} &
\includegraphics[width=0.163\textwidth]{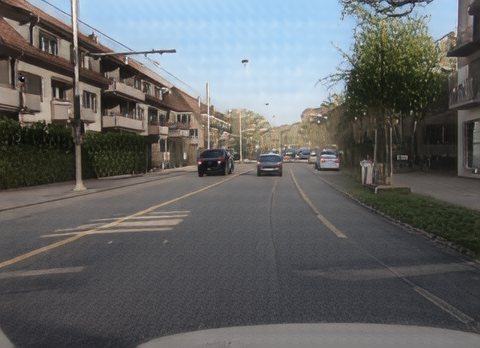} &
\includegraphics[width=0.163\textwidth]{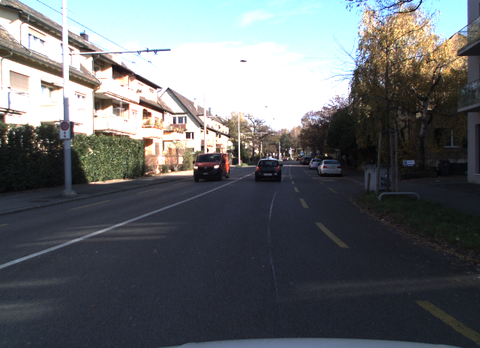} \vspace{-2.5pt} \\ 

\includegraphics[width=0.163\textwidth]{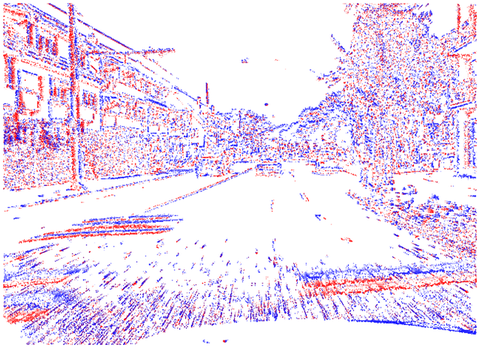} &
\includegraphics[width=0.163\textwidth]{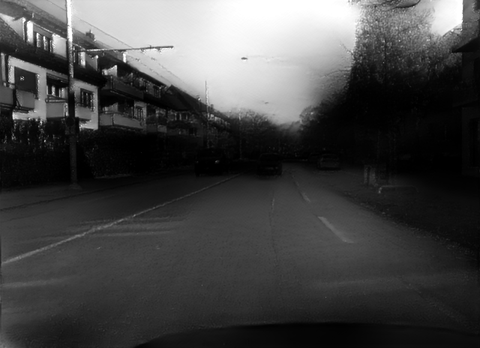} &
\includegraphics[width=0.163\textwidth]{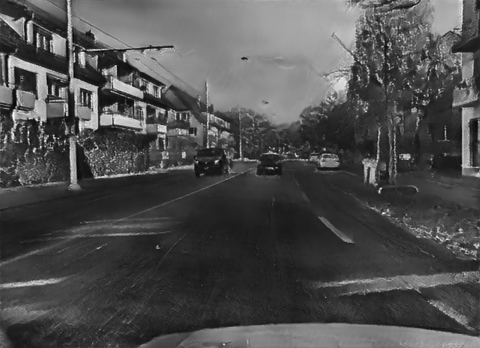} &
\includegraphics[width=0.163\textwidth]{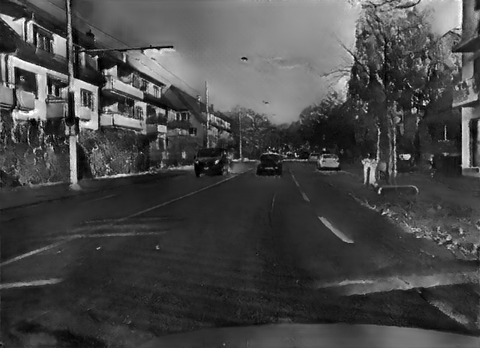} &
\includegraphics[width=0.163\textwidth]{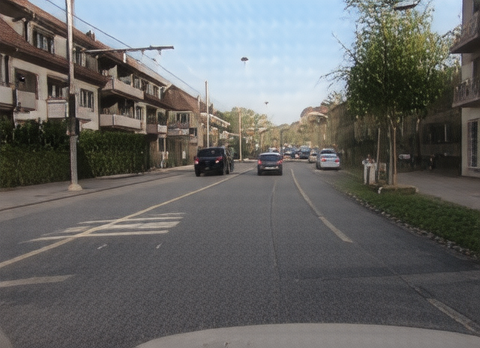} &
\includegraphics[width=0.163\textwidth]{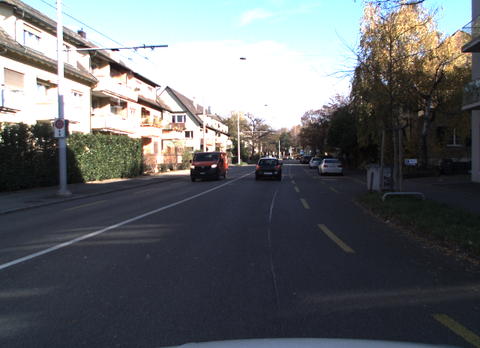} \vspace{-2.5pt} \\ 

\includegraphics[width=0.163\textwidth]{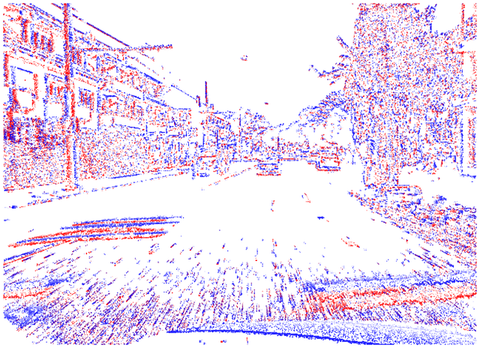} &
\includegraphics[width=0.163\textwidth]{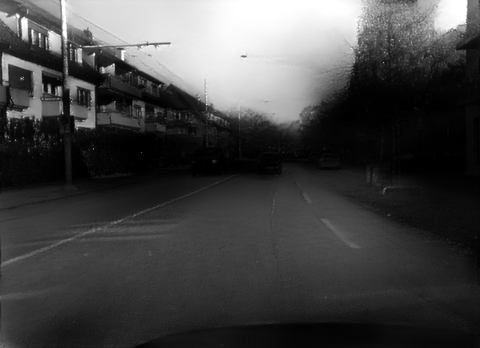} &
\includegraphics[width=0.163\textwidth]{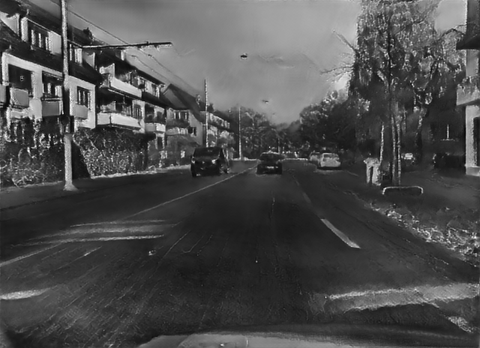} &
\includegraphics[width=0.163\textwidth]{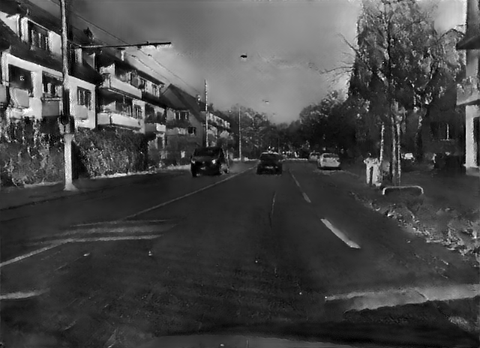} &
\includegraphics[width=0.163\textwidth]{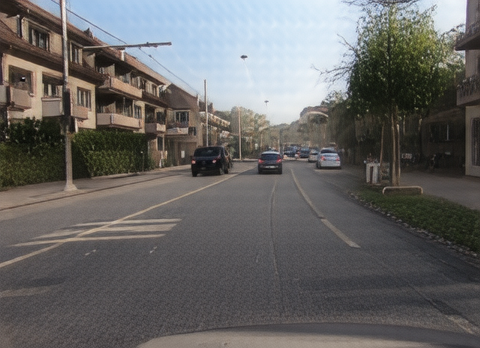} &
\hspace{-5pt}
\includegraphics[width=0.163\textwidth]{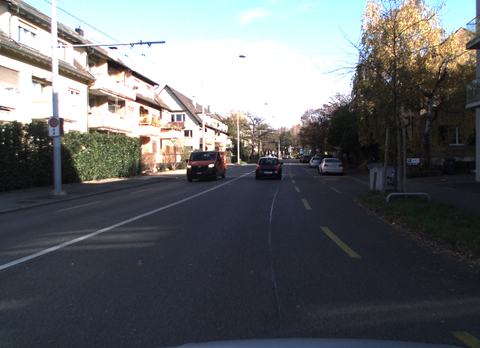} \\
\multicolumn{1}{c}{\scriptsize Events} & 
\multicolumn{1}{c}{\scriptsize E2VID~\cite{rebecq2019events,rebecq2019high}} & 
\multicolumn{1}{c}{\scriptsize ET-NET~\cite{ET-Net}} & 
\multicolumn{1}{c}{\scriptsize HyperE2VID~\cite{ercan2024hypere2vid}} & 
\multicolumn{1}{c}{\scriptsize \textbf{EvDiff (Ours)}} & 
\multicolumn{1}{c}{\scriptsize GT} \\

\end{tabular}
\caption{\textbf{Visual comparison on BS-ERGB~\cite{tulyakov2022time} and DSEC~\cite{dsec} datasets}. Our EvDiff produces higher-quality chromatic frames.}
\label{fig:multi_comparison}
\end{figure}
\begin{figure}[!t] 
\centering
\setlength{\tabcolsep}{0.5pt} 

\begin{tabular}{cccccc}

\includegraphics[width=0.163\textwidth]{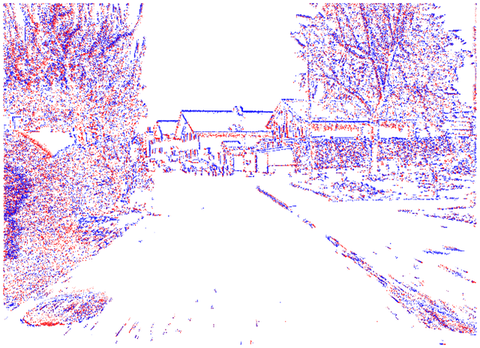} &
\includegraphics[width=0.163\textwidth]{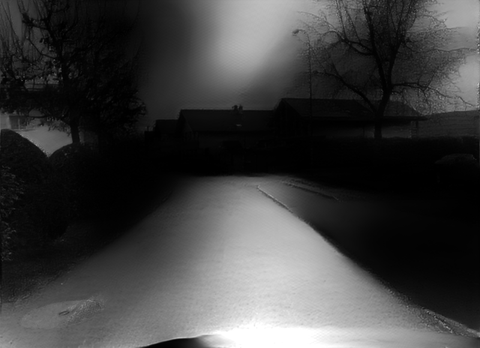} &
\includegraphics[width=0.163\textwidth]{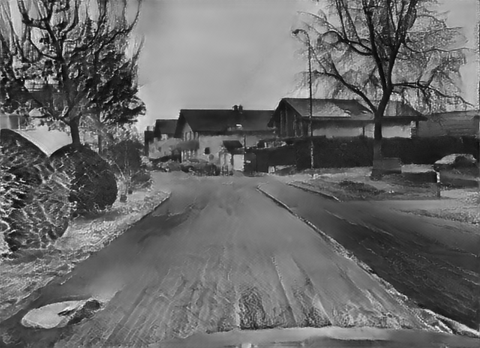} &
\includegraphics[width=0.163\textwidth]{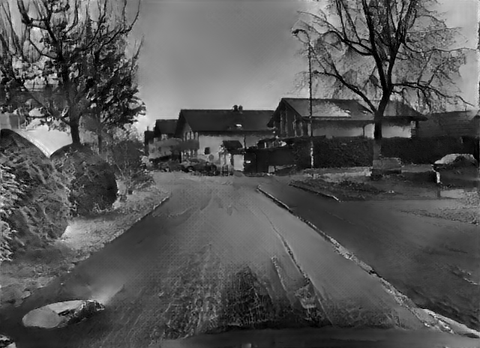} &
\includegraphics[width=0.163\textwidth]{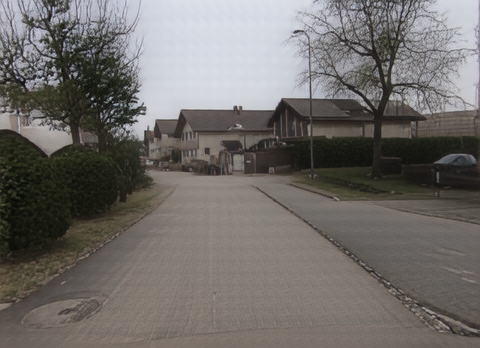} &
\includegraphics[width=0.163\textwidth]{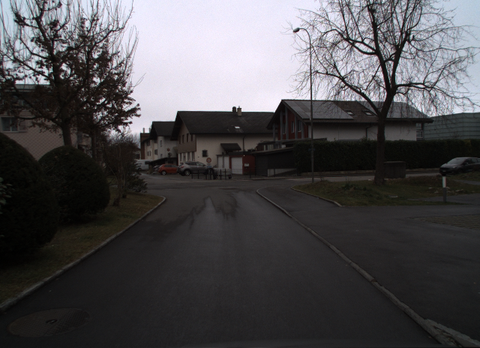} \vspace{-2.5pt} \\ 

\includegraphics[width=0.163\textwidth]{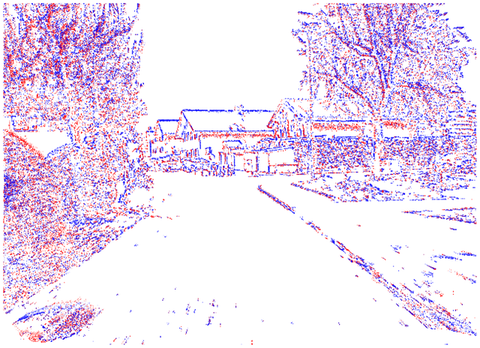} &
\includegraphics[width=0.163\textwidth]{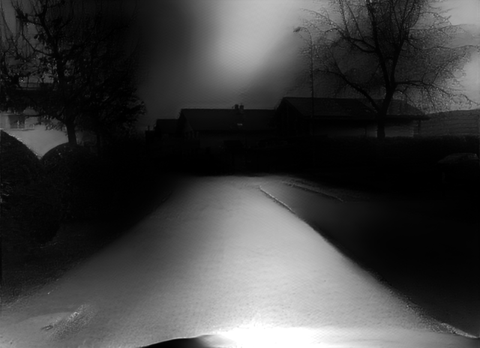} &
\includegraphics[width=0.163\textwidth]{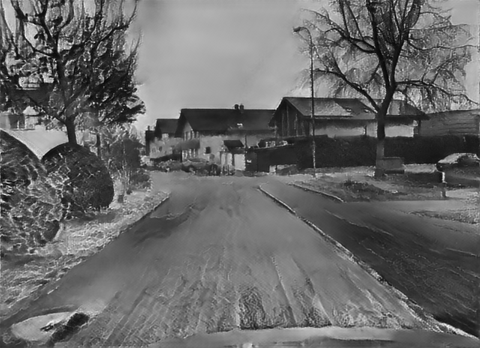} &
\includegraphics[width=0.163\textwidth]{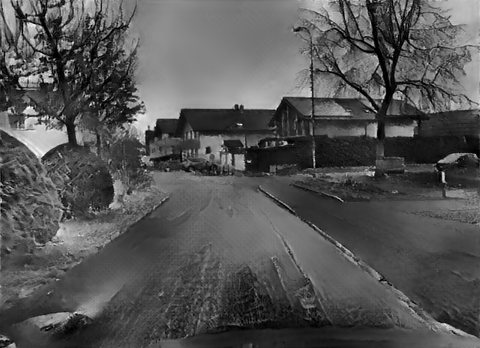} &
\includegraphics[width=0.163\textwidth]{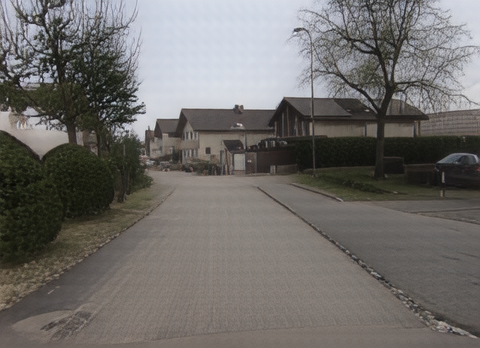} &
\includegraphics[width=0.163\textwidth]{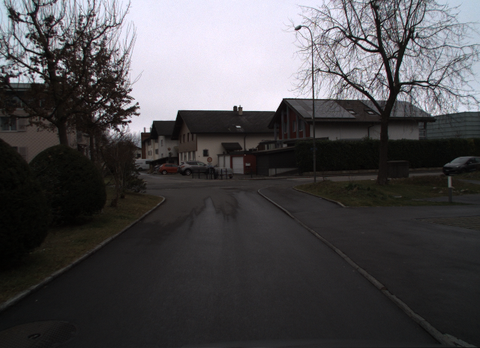} \vspace{-2.5pt} \\ 

\includegraphics[width=0.163\textwidth]{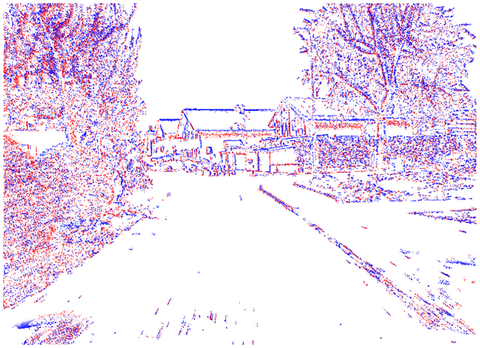} &
\includegraphics[width=0.163\textwidth]{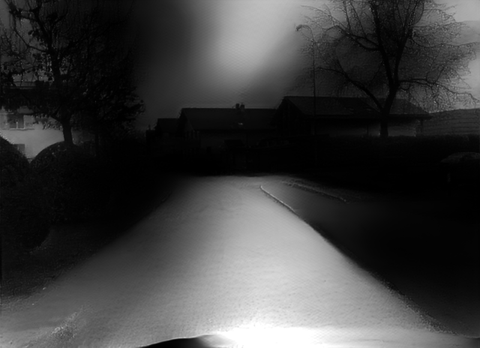} &
\includegraphics[width=0.163\textwidth]{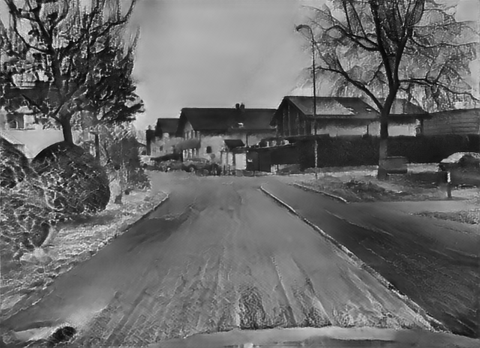} &
\includegraphics[width=0.163\textwidth]{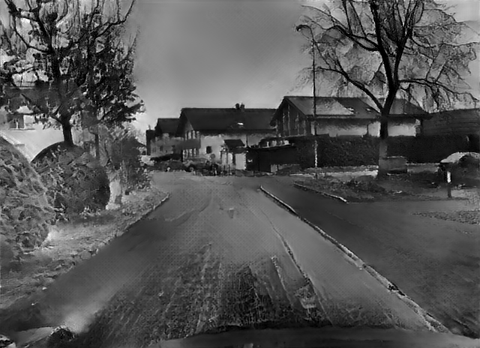} &
\includegraphics[width=0.163\textwidth]{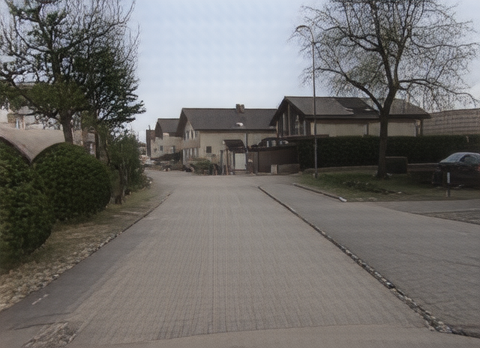} &
\includegraphics[width=0.163\textwidth]{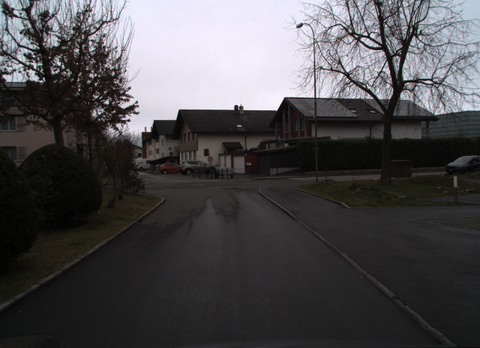} \vspace{-2.5pt} \\ 

\includegraphics[width=0.163\textwidth]{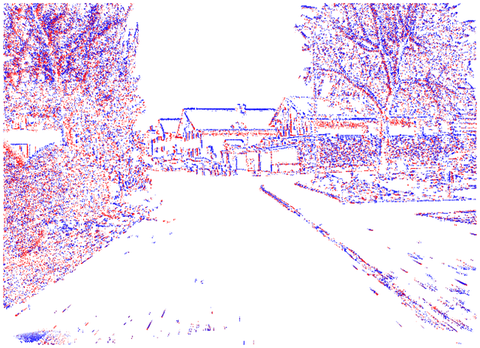} &
\includegraphics[width=0.163\textwidth]{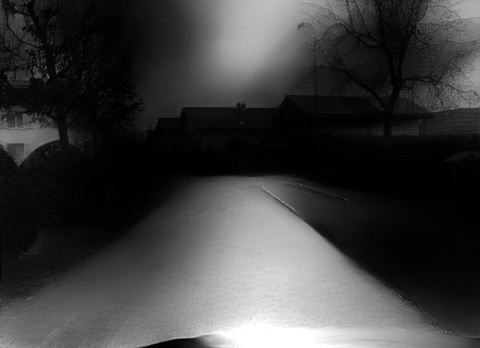} &
\includegraphics[width=0.163\textwidth]{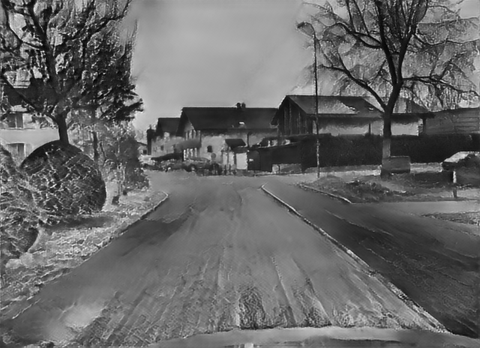} &
\includegraphics[width=0.163\textwidth]{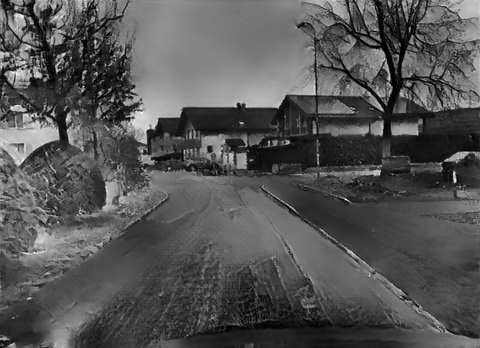} &
\includegraphics[width=0.163\textwidth]{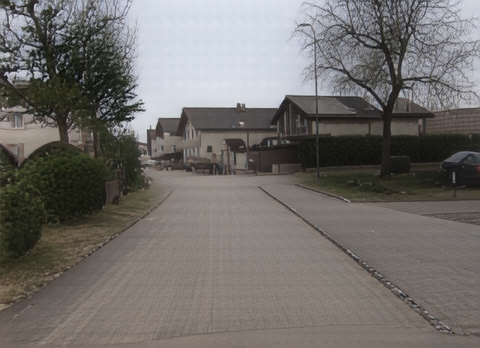} &
\includegraphics[width=0.163\textwidth]{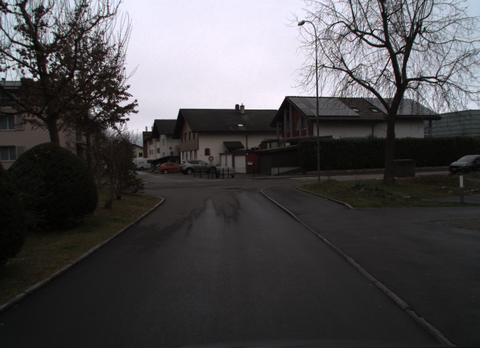} \vspace{-2.5pt} \\ 

\includegraphics[width=0.163\textwidth]{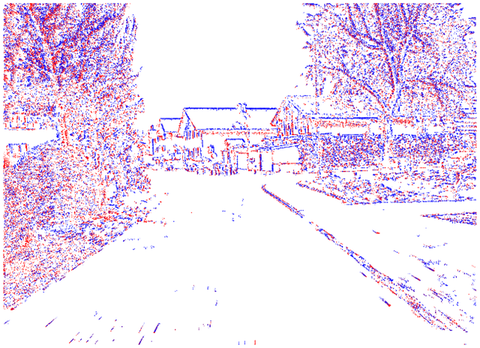} &
\includegraphics[width=0.163\textwidth]{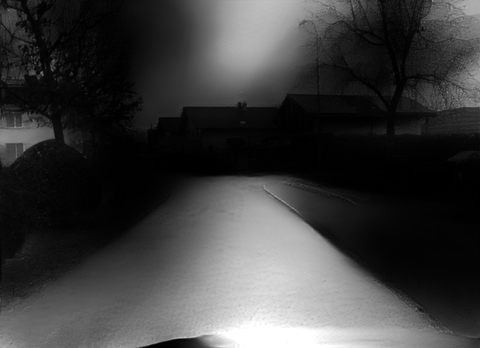} &
\includegraphics[width=0.163\textwidth]{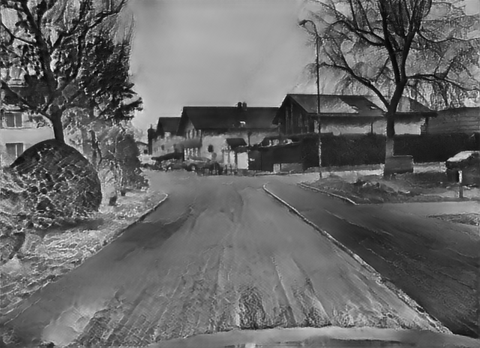} &
\includegraphics[width=0.163\textwidth]{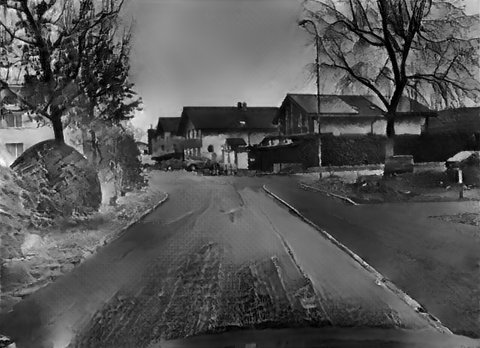} &
\includegraphics[width=0.163\textwidth]{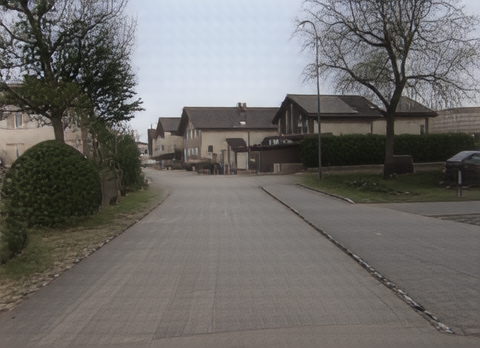} &
\includegraphics[width=0.163\textwidth]{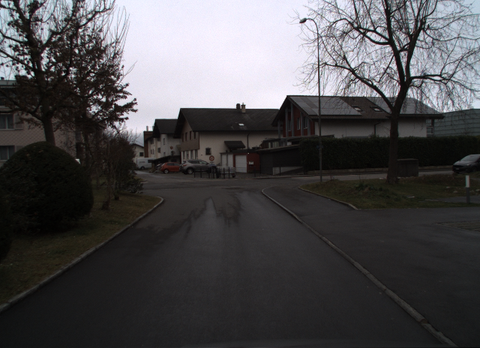} \vspace{-2.5pt} \\ 

\includegraphics[width=0.163\textwidth]{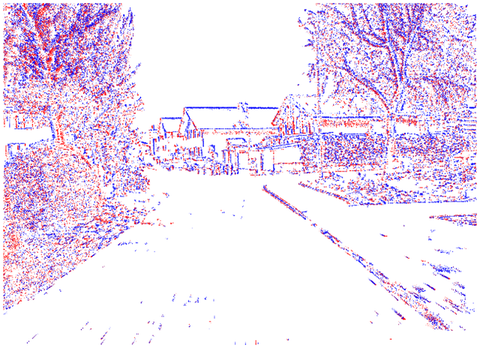} &
\includegraphics[width=0.163\textwidth]{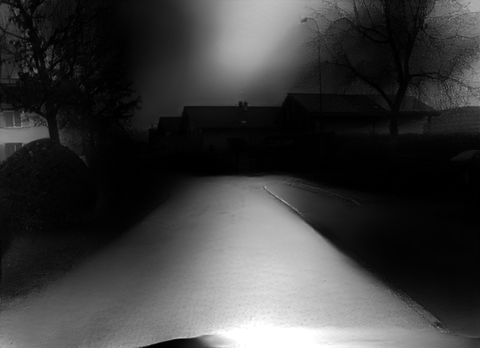} &
\includegraphics[width=0.163\textwidth]{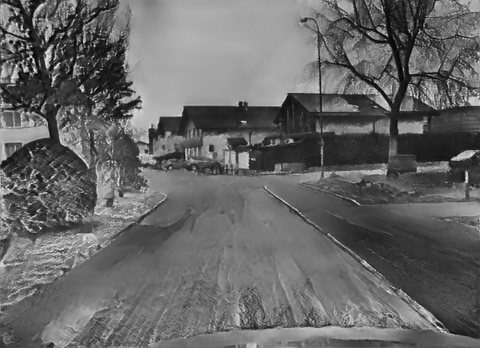} &
\includegraphics[width=0.163\textwidth]{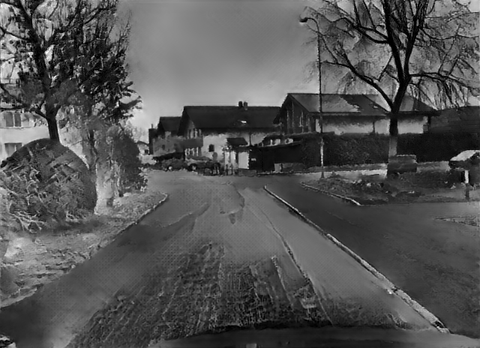} &
\includegraphics[width=0.163\textwidth]{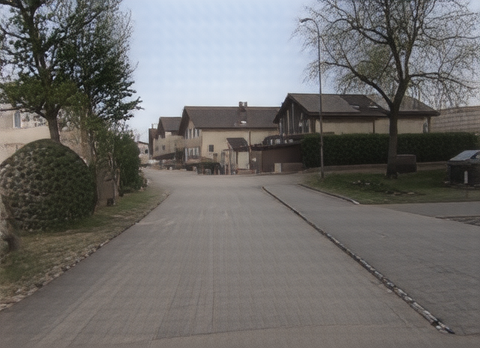} &
\hspace{-5pt}
\includegraphics[width=0.163\textwidth]{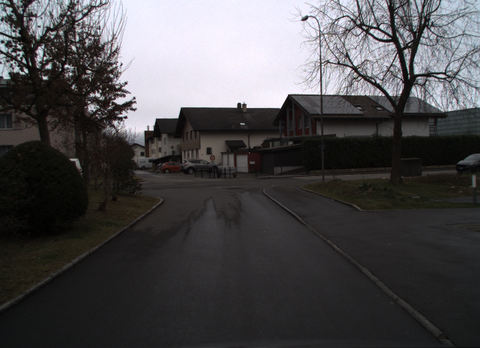} \\

\multicolumn{1}{c}{\scriptsize Events} & 
\multicolumn{1}{c}{\scriptsize E2VID~\cite{rebecq2019events,rebecq2019high}} & 
\multicolumn{1}{c}{\scriptsize ET-NET~\cite{ET-Net}} & 
\multicolumn{1}{c}{\scriptsize HyperE2VID~\cite{ercan2024hypere2vid}} & 
\multicolumn{1}{c}{\scriptsize \textbf{EvDiff (Ours)}} & 
\multicolumn{1}{c}{\scriptsize GT} \\

\end{tabular}
\caption{\textbf{Visual comparison on DSEC~\cite{dsec} datasets}. Our EvDiff produces higher-quality chromatic frames.}
\label{fig:more_multi_comparison}
\end{figure}

\begin{table}[!t]
\centering
\small
\renewcommand\arraystretch{1.05}
\caption{\textbf{Comparison of parameters and FLOPs} between our EvDiff and ControlNet at different resolutions.}
\vspace{-9pt}
\resizebox{1.0\textwidth}{!}{
\setlength{\tabcolsep}{9mm}
{
\begin{tabular}{l||cccc}
\bottomrule[0.12em]
\rowcolor{tableHeadGray}
\textbf{Methods} & 
\textbf{Ours} & 
\makecell{\textbf{ControlNet}\\\textbf{in 10 steps}} & 
\makecell{\textbf{ControlNet}\\\textbf{in 20 steps}} &
\makecell{\textbf{ControlNet}\\\textbf{in 40 steps}} \\ \hline \hline
\#Params & 2.19G & 8.23G & 8.23G & 8.23G \\ \arrayrulecolor{gray}\hdashline\arrayrulecolor{black}
\makecell[l]{Flops 512$\times$512} & 2.18T & 26.62T & 49.93T & 96.56T \\ \arrayrulecolor{gray}\hdashline\arrayrulecolor{black}
\makecell[l]{Flops 1024$\times$1024} & 9.56T & 87.15T & 163.55T & 316.36T \\ 
\hline
\end{tabular}
}
}
\vspace{-3pt}
\label{tab:params_flops}
\end{table}


\begin{figure*}[t]
\centering
\setlength{\tabcolsep}{0.5pt} 
\renewcommand{\arraystretch}{0.8} 
\begin{tabular}{cccccc}
\includegraphics[width=0.163\textwidth]{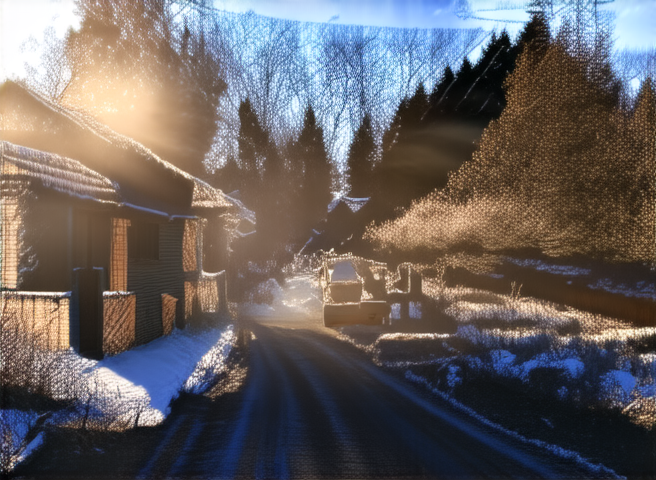} &
\includegraphics[width=0.163\textwidth]{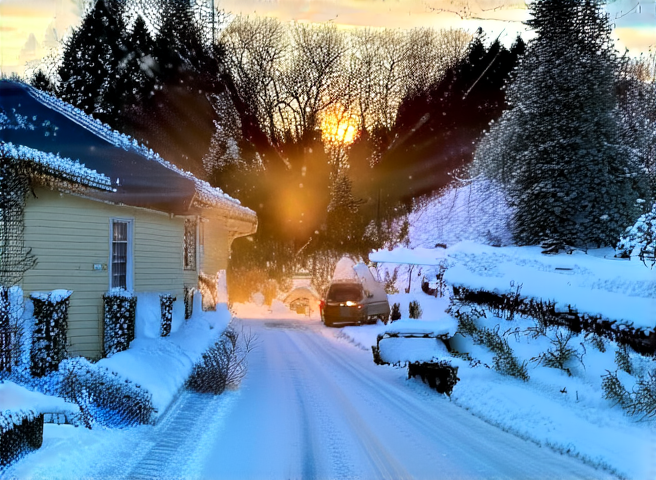} &
\includegraphics[width=0.163\textwidth]{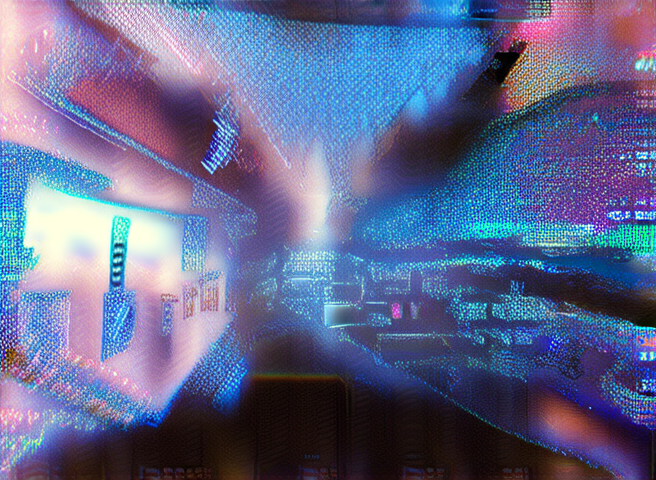} &
\includegraphics[width=0.163\textwidth]{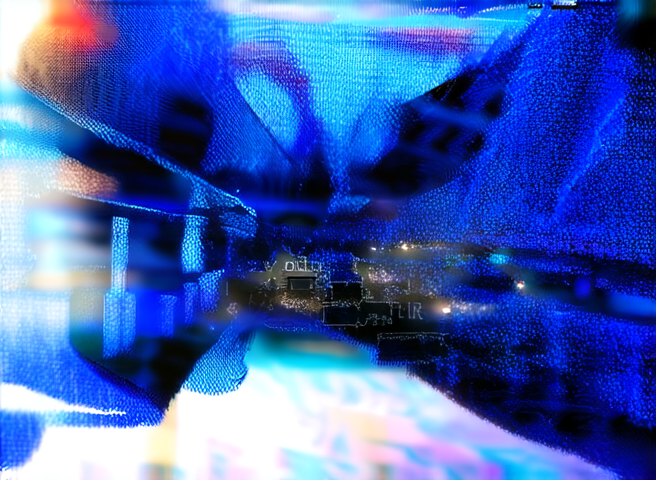} &
\includegraphics[width=0.163\textwidth]{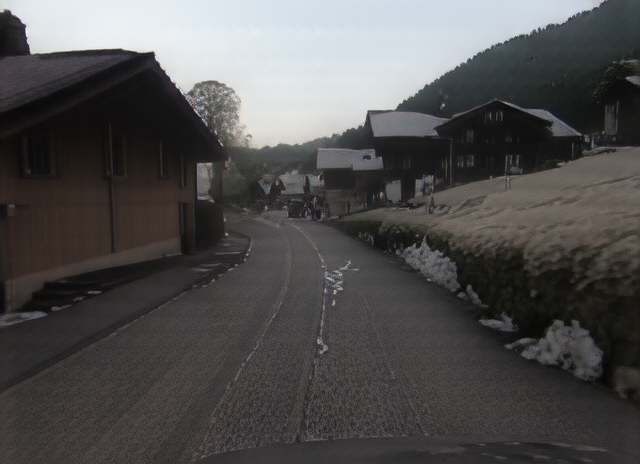} &
\includegraphics[width=0.16\textwidth]{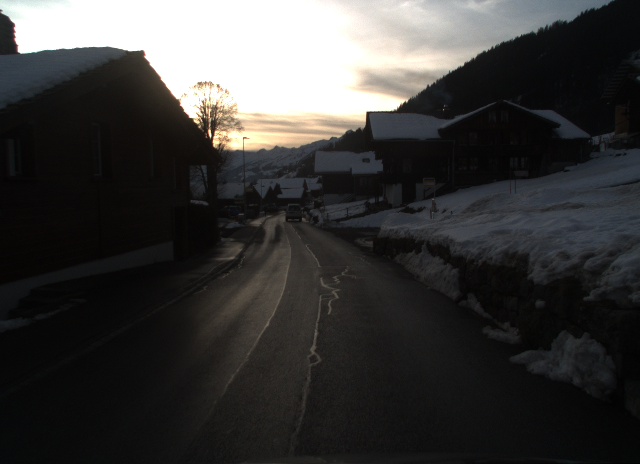}  \\[-2pt]
\includegraphics[width=0.163\textwidth]{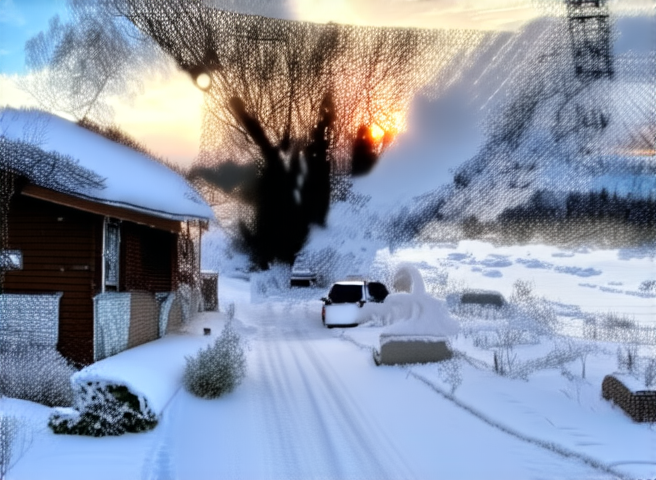} &
\includegraphics[width=0.163\textwidth]{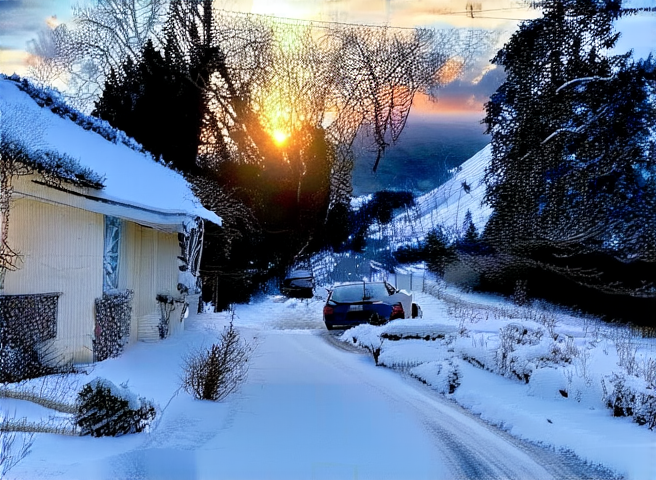} &
\includegraphics[width=0.163\textwidth]{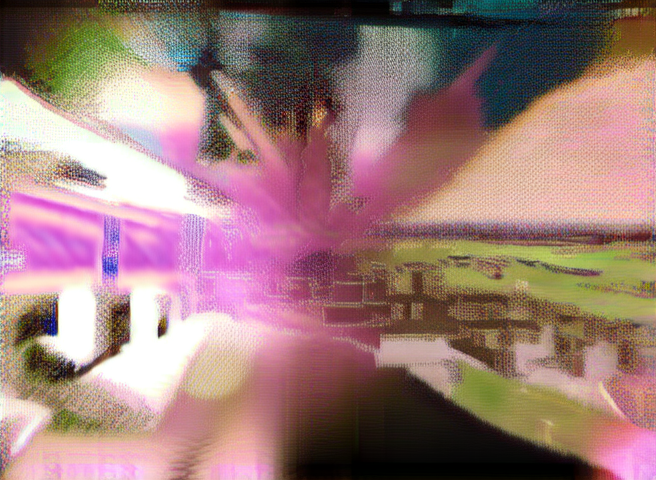}  &
\includegraphics[width=0.163\textwidth]{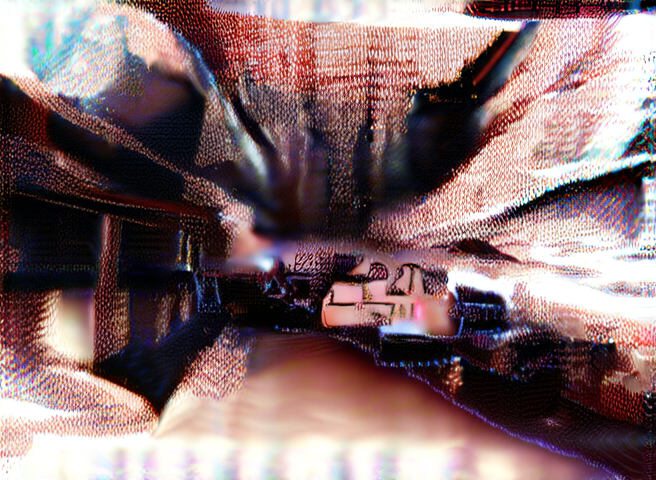}&
\includegraphics[width=0.163\textwidth]{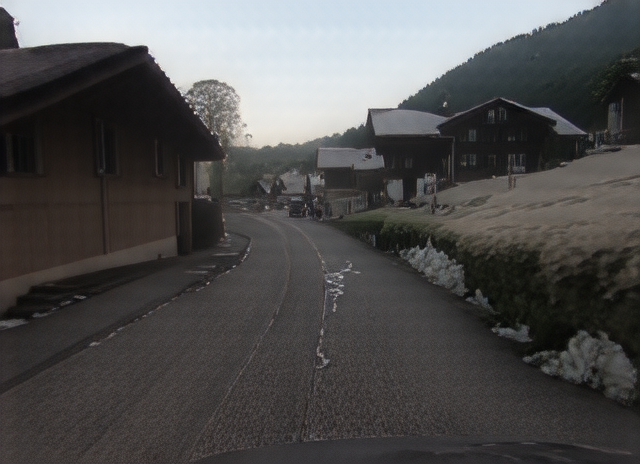} &
\includegraphics[width=0.163\textwidth]{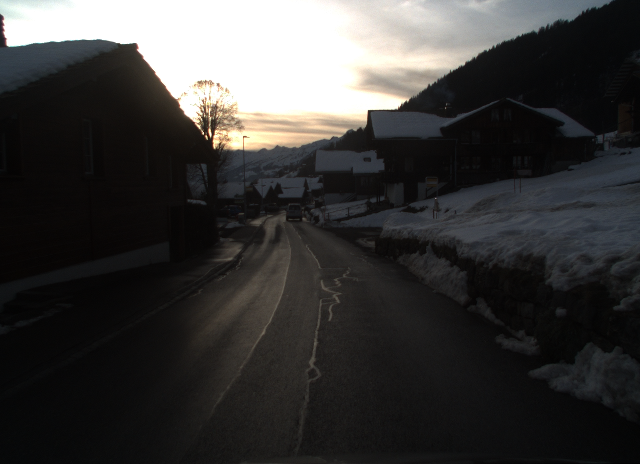} \\[-2pt]
\multicolumn{1}{c}{\makecell{\scriptsize GT Prompt\\in 10 Steps}} &
\multicolumn{1}{c}{\makecell{\scriptsize GT Prompt\\in 40 Steps}} &
\multicolumn{1}{c}{\makecell{\scriptsize Fixed Prompt\\in 10 Steps}} &
\multicolumn{1}{c}{\makecell{\scriptsize Fixed Prompt\\in 40 Steps}} &
\multicolumn{1}{c}{\makecell{\scriptsize Fixed Prompt\\\textbf{Ours in 1 step}}} &
\multicolumn{1}{c}{\makecell{\scriptsize GT}} \\
\end{tabular}
\caption{\textbf{Visual comparison agains multi-steps ControlNet-based counterparts}. Left four columns: Results from ControlNet-based methods. ``GT Prompt'': Prompts are produced from GT images with RAM~\cite{ram}. Our EvDiff produces more faithful results.}
\label{fig:controlnet_comparison}
\end{figure*}

\begin{table*}[t]
\centering
\small
\setlength{\dashlinedash}{2pt}
\setlength{\dashlinegap}{1.5pt}
\setlength{\arrayrulewidth}{0.3pt}

\setlength{\tabcolsep}{10pt}
\caption{\textbf{Ablation study on different variants of our method,} with the best results highlighted in \best{red} and the second best in \second{blue}.}
\vspace{-8pt}
\resizebox{1.0\textwidth}{!}
{
\renewcommand\arraystretch{1.1}

    \begin{tabular}{l||ccccc||ccccc}
    \bottomrule[0.12em]
    \rowcolor{tableHeadGray} 
    \textbf{Method} & \textbf{MSE$\downarrow$} & \textbf{SSIM$\uparrow$} & \textbf{LPIPS$\downarrow$} & \textbf{FID$\downarrow$} & \textbf{FVD$\downarrow$} & \textbf{MSE$\downarrow$} & \textbf{SSIM$\uparrow$} & \textbf{LPIPS$\downarrow$} & \textbf{FID$\downarrow$} & \textbf{FVD$\downarrow$} \\ \hline \hline
    & \multicolumn{5}{c||}{\textbf{BSERGB}} & \multicolumn{5}{c}{\textbf{DSEC}} \\ \hline
    \rowcolor{gray!8}
   \textit{w./o.} Stage 1 training & 0.0726 & 0.2255 & 0.4906 & 374 & 2254 & 0.0630 & 0.3141 & 0.4817 & 256 & 2063 \\    
    \textit{w./o.} E2VID-Style Degradation Model & 0.0850 & 0.1972 & 0.5267 & 396 & 3683 & 0.0918 & 0.2413 & 0.5286 & 450 & 3711 \\
    \rowcolor{gray!8}
    \textit{w./o.} EGM Degradation & 0.0738 & 0.2301 & 0.5000 & 276 & 1683 & 0.0702 & 0.2973 & 0.4937 & 260 & 1505 \\
    \textit{w./o.} SII Degradation & 0.0667 & 0.2478 & 0.4743 & 277 & 1226 & 0.0734 & 0.3017 & 0.4897 & 238 & 1314 \\
    \rowcolor{gray!8}
    \textit{w./o.} AD Degradation & 0.0747 & 0.223 & 0.5059 & 333 & 2324 & 0.0861 & 0.2774 & 0.5150 & 333 & 2075 \\
    \textit{w./o.} Surrogate Distilling & 0.0660 & 0.2822 & 0.4373 & 162 & 957 & 0.0778 & 0.2818 & 0.4750 & 164 & 1539 \\
    \rowcolor{gray!8}
    \textit{w./o.} Stage 3 finetune & 0.0763 & 0.2647 & 0.5740 & 224 & 1726 & 0.0968 & 0.2421 & 0.6546 & 189 & 1503 \\ 
    
    \arrayrulecolor{gray}\hdashline\arrayrulecolor{black}
    
    \textit{w./o.} ETF & 0.0586 & 0.2966 & 0.4234 & 159 & 1299 & 0.0605 & 0.3265 & 0.4482 & 152 & 1449 \\
    \rowcolor{gray!8}
    \textit{w.} ConvGRU & 0.0500 & 0.3302 & \second{0.4147} & 165 & \second{909} & \second{0.0561} & 0.3383 & \second{0.4337} & \best{127} & \best{985} \\
    \textit{w.} ConvLSTM & \second{0.0496} & \second{0.3363} & 0.4156 & \second{160} & \best{907} & 0.0620 & 0.3312 & 0.4403 & \best{127} & \second{1039} \\
    \rowcolor{gray!8}
    \textit{w.} E2VID~+~VAE-Encoder & 0.0729 & 0.3322 & 0.4827 & 231 & 1551 & 0.0580 & \best{0.3849} & 0.4691 & 184 & 1368 \\
    \arrayrulecolor{gray}\hdashline\arrayrulecolor{black}
    
    \rowcolor{gray!8}
    Final Model & \best{0.0463} & \best{0.3394} & \best{0.4023} & \best{148} & 984 & \best{0.0476} & \second{0.3677} & \best{0.4226} & 129 & 1491 \\
    \hline
    \end{tabular}

}
\label{tab:ablation}
\vspace{-6pt}
\end{table*}

\section{Experiment}

\subsection{Experimental Settings}

\noindent\textbf{Training Set.}
We adopt the Place365 dataset~\cite{places365} as our training set for stage 1, which contains 1,800,000 HQ images. 
For stage 2 \& 3, we use the high-frame-rate videos from the REDS dataset~\cite{nah2019ntire}, which contains 240 seq and 500 images each sequence. v2e~\cite{hu2021v2e} simulator is used to produce synthetic events.

\noindent\textbf{Implementation Details.}
EvDiff is trained with the 3-stage Surrogate Training Pipeline. All models are fine-tuned using LoRA~\cite{hu2022lora} with rank~64 and optimized by AdamW~\cite{loshchilov2017decoupled} 
with an initial learning rate of $5\times10^{-6}$ ($\beta_1=0.9$, $\beta_2=0.999$). Training is conducted on a single NVIDIA H200 GPU. Stage~1 trains the diffusion model for 180k iterations with a batch size of 10. Stage~2 performs Surrogate Distillation for 12k iterations (batch size 1, sequence length 40), followed by Stage~3 joint fine-tuning for 12k iterations (batch size 1, sequence length 30). During Stages~2 and~3, we apply online degradation to dynamically corrupt simulated event streams (e.g., event merging, dropping, and polarity removal) to better mimic real sensor behavior. More details are provided in the \textit{supp}.


\noindent\textbf{Evaluation.}
We evaluate our method on two real-world datasets: BS-ERGB~\cite{tulyakov2022time} and DSEC~\cite{dsec} datasets. Both of them contain chromatic RGB frames and monochromatic events. We follow the official test split for DSEC and the split proposed in EVReal~\cite{evreal} for BS-ERGB, respectively. HQF~\cite{E2VID+} and MVSEC~\cite{zhu2018multivehicle} are excluded due to their suboptimal quality and the lack of color information.
We report both standard fidelity metrics (MSE, SSIM) and perceptual metrics (LPIPS, FID, FVD).
For fair comparison with existing methods (\eg, ETNet~\cite{ET-Net}, HyperE2VID~\cite{ercan2024hypere2vid}), the results from our method are converted to grayscale images before computing MSE, SSIM and LPIPS metrics.

\subsection{Comparisons with State-of-the-Art Methods}
\label{subsec:comparisons}

We compare our method with representative event-based video reconstruction approaches, including E2VID~\cite{rebecq2019events}, FireNet~\cite{2020FireNet}, E2VID+~\cite{E2VID+}, FireNet+~\cite{E2VID+}, SPADE-E2VID~\cite{Cadena2021Spade-E2VID}, ETNet~\cite{ET-Net}, SSL-E2VID~\cite{SSL-E2VID}, and HyperE2VID~\cite{ercan2024hypere2vid}. For all methods official weights are used. The numerical results on reference metrics are shown in Table~\ref{tab:results}.

Considering an improvement of $32.2\% / 8.5\% / 37.8\%$ and $20.8\% / 14.0\% / 44.9\%$ in MSE/LPIPS/FID compared to second best method on BS-ERGB and DSEC, respectively, the proposed EvDiff delivers overall state-of-the-art performance, showing advantages in perceptual (LPIPS) and visual realism (FID, FVD), while preserving solid fidelity (MSE, SSIM).
Specifically, traditional E2VID and its variants aim to improve robustness to complex event inputs for enhancing fidelity, but their generated videos still show poor perceptual and generative quality due to the lack of priors from large dataset pretraining. 
This is also evident from the qualitative results in Figure~\ref{fig:multi_comparison}, where the E2VID family exhibits severe visually unpleasant artifacts across the entire image, and the generated details lack realism. 
In contrast, benefiting from DiT training, which converts large-scale high-quality datasets into diffusion priors that can handle E2VID-style degradation artifacts, and the temporal-based EvEncoder designed for events characteristics in surrogate distillation, our EvDiff achieves significant advantages in perceptual and generative quality evaluations. 
Furthermore, our EvDiff is the \textit{only} method that generates chromatic video from monochromatic events. In Fig.~\ref{fig:multi_comparison}, our results retain the cloud shape, while the GT image is over-exposed. This shows that EvDiff offers higher effective dynamic range thanks to the HDR properties of event cameras.

\subsection{Comparisons with the ControlNet-Based Counterpart}
\label{subsec:comparisons_controlnet}
As described in \S\ref{subsec:architecture}, a more intuitive idea for using diffusion models for event-based video reconstruction is a ControlNet-based multi-step diffusion method. We use implementation from HuggingFace~\cite{diffusers_controlnet_sd3_2024} and keep the SD3 base model since there is no open-sourced diffusion-based E2VID method.
As shown in Fig.~\ref{fig:controlnet_comparison}, our EvDiff produces results that are more temporally consistent and more faithful to the real scene, \ie, with higher \textit{fidelity}.
Moreover, when provided with the same scene-agnostic prompts (``Fixed Prompt'') used in our model, ControlNet largely fails to achieve effective reconstruction. This is crucial for video reconstruction, where events are the only input and human prior knowledge is typically unavailable.
Finally, in the single-frame case, as shown in Table~\ref{tab:params_flops}, our EvDiff contains fewer parameters and lower computational cost, not to mention that in the multi-frame setting, the computational complexity only scales linearly with the frame number $T$. 


\subsection{Ablation Study}
\label{subsec:ablation}

We conduct ablation studies on both the BS-ERGB and DSEC datasets to verify the contribution of each key component, as shown in Table~\ref{tab:ablation}. For the proposed Surrogate Training Pipeline, removing Stage~1 leads to a 56.8\% performance drop, even when initialized with pretrained SD3 weights, highlighting the necessity of large-scale dataset training. 
Excluding our E2VID-Style Degradation Model still results in a 83.6\% performance decrease, demonstrating its importance in bridging the domain gap. We further analyze the proposed Degradation Model by ablating its internal components. When one component is excluded (either EGM, SII or AD), the performance remains inferior to the full model. Surrogate Distillation boosts performance by 29.8\%, as it aligns the EvEncoder with the VAE latent space and ensures a clean, stable input for the diffusion refiner.
Stage~3 joint fine-tuning is also critical. Without it, imperfections in the EvEncoder latent representations degrade the final reconstruction. Joint fine-tuning allows the DiT refiner to adapt to these imperfections and recover high-quality results.

Regarding the model architecture, the ETF module improves MSE and FVD by 20.9\% and 24.2\%, respectively. Replacing ETF with ConvLSTM or ConvGRU (with comparable model size) consistently degrades performance, confirming the effectiveness of the proposed temporal fusion design. Replacing our EvEncoder with the ``E2VID + VAE-Encoder'' baseline further degrades perceptual quality while making the pipeline 2.5$\times$ larger and slower (84\,ms vs.\ 63\,ms), demonstrating that our EvEncoder achieves a better balance between efficiency and performance.
\section{Conclusion}


Event-to-video reconstruction has long been constrained by limited training data, weak generalization, and low restoration quality. By decomposing the problem and introducing the Surrogate Training Pipeline, EvDiff transfers the power of large pretrained generative models to event-based video reconstruction and enables training with large-scale image datasets. The resulting one-step EvDiff produces high-quality, photorealistic, and faithful reconstructions from event streams while remaining competitive with ControlNet-based approaches. We hope this work helps bridge event-based vision and modern generative models and inspires future research on temporally consistent event-to-video reconstruction.

\section*{Acknowledgements}
This research was supported by Zhejiang Provincial Natural Science Foundation of China under Grant No. LZ24F050003. It was also partially funded by the Ministry of Education and Science of Bulgaria, as part of the Bulgarian National Roadmap for Research Infrastructure (support for INSAIT) and the European Union's Horizon Europe -- the Framework Programme for Research and Innovation, under grant agreement 101168521.


%
%
\bibliographystyle{splncs04}
\bibliography{main}

\end{document}